\documentclass[11pt]{scaleai-paper}

\usepackage{amsmath}
\usepackage{amssymb}
\usepackage{booktabs}
\usepackage{multirow}
\usepackage{array}
\usepackage{tabularx}
\usepackage{longtable}
\usepackage{graphicx}
\usepackage{bm}
\usepackage{courier}
\usepackage{listings}
\usepackage{float}
\usepackage{wrapfig}
\usepackage{placeins}
\usepackage{pdflscape}
\usepackage{tikz}
\usetikzlibrary{arrows.meta,positioning,calc,fit,backgrounds,shapes.geometric}
\usepackage{pgfplots}
\usepgfplotslibrary{groupplots}
\pgfplotsset{compat=1.18}
\usepackage[square,numbers,sort&compress]{natbib}
\usepackage{xspace}
\usepackage{url}
\usepackage[colorlinks=true,linkcolor=scaleLink,citecolor=scaleLink,urlcolor=scaleLink]{hyperref}
\hypersetup{
  pdftitle={SteerDuplex: Steerable Duplex Speech Dialogue Models},
  pdfauthor={Utkarsh Tyagi, Ramaneswaran Selvakumar, Advait Gosai, Sonal Kumar, Nikhil Barhate, Isabell Sagar, Steven Li, Miheer Bavare, Daniel Quigley, Fabiola Tapia Carrillo, Jose M Patron E, Diego Macias Gutierrez, Paul Song, Ramani Duraiswami, Dinesh Manocha, Yunzhong He}
}
\definecolor{hackblue}{HTML}{2F80ED}
\definecolor{hackred}{HTML}{D55E00}
\definecolor{hackgreen}{HTML}{009E73}
\definecolor{hackgold}{HTML}{E69F00}
\definecolor{hackpurple}{HTML}{6A51A3}
\definecolor{hackgray}{HTML}{6B7280}
\definecolor{moshicol}{HTML}{8C8C99}
\definecolor{personacol}{HTML}{5FA5C7}
\definecolor{sftcol}{HTML}{E69F00}
\definecolor{oursalmcol}{HTML}{E78A6B}
\definecolor{paretocol}{HTML}{B8B8C0}
\definecolor{qualblue}{HTML}{2F80ED}
\definecolor{qualgreen}{HTML}{009E73}
\definecolor{qualgold}{HTML}{E69F00}
\definecolor{qualpurple}{HTML}{6A51A3}
\lstdefinestyle{steerprompt}{
  basicstyle=\ttfamily\footnotesize,
  columns=fullflexible,
  keepspaces=true,
  breaklines=true,
  breakatwhitespace=true,
  showstringspaces=false,
  aboveskip=0pt,
  belowskip=0pt
}
\newtcblisting{promptlisting}{
  enhanced,
  listing only,
  colback=scalePanel,
  colframe=scaleMediumGray,
  boxrule=0.5pt,
  arc=2pt,
  left=7pt,
  right=7pt,
  top=6pt,
  bottom=6pt,
  before skip=6pt plus 2pt minus 1pt,
  after skip=7pt plus 2pt minus 1pt,
  listing options={style=steerprompt}
}
\newcommand{\emailicon}{%
  \tikz[baseline=-0.68ex,x=1.55ex,y=1.55ex]{%
    \fill[black,rounded corners=0.08ex] (0,0) rectangle (1.35,0.9);%
    \draw[white,line width=0.075ex] (0.06,0.80) -- (0.675,0.34) -- (1.29,0.80);%
  }%
}

\papertype{Scale AI Research}
\contact{\emailicon\ \href{mailto:utkarsh.tyagi@scale.com}{\textcolor{black}{\texttt{utkarsh.tyagi@scale.com}}}\enspace\textbar\enspace\url{https://github.com/Utkarsh4430/SteerDuplex}}

\title{\textsc{SteerDuplex}: Steerable Duplex Speech Dialogue Models}

\author[1*$\dagger$]{Utkarsh Tyagi}
\author[2*]{Ramaneswaran Selvakumar}
\author[1]{Advait Gosai}
\author[2]{Sonal Kumar}
\author[1]{Nikhil Barhate}
\author[1]{Isabell Sagar}
\author[1]{Steven Li}
\author[1]{Miheer Bavare}
\author[1]{Daniel Quigley}
\author[1]{Fabiola Tapia Carrillo}
\author[1]{Jose M Patron E}
\author[1]{Diego Macías Gutiérrez}
\author[1]{Paul Song}
\author[2]{Ramani Duraiswami}
\author[2]{Dinesh Manocha}
\author[1]{Yunzhong He}
\affil[1]{Scale AI}
\affil[2]{University of Maryland}
\affil[]{\textsuperscript{*}Equal contribution. \textsuperscript{$\dagger$}Project lead.}

\begin{document}
\maketitle
\begingroup
\linespread{1.09}\selectfont
\begin{abstract}
Full-duplex spoken dialogue models support low-latency turn taking, interruption handling, and backchanneling, yet a key capability remains underexplored: \emph{steerability}, the ability to reliably shift conversational behavior along attributes such as tone, persona, speaking rate, and voice style in response to user instructions. We introduce a taxonomy of text- and audio-based steerability that identifies substantial gaps in current full-duplex models. To address this gap, we introduce \textsc{SteerDuplex}, a Moshi-based full-duplex speech model fine-tuned on natural conversations and synthetic dialogues targeting instruction following, vocal delivery, reasoning, and duplex interaction. We further apply two-stage reinforcement learning (RL) with hybrid rewards, combining verifiable interaction checks and judge-based semantic feedback to improve timing and response continuity. To evaluate full-duplex spoken steerability, we introduce \textsc{SteerBench}, a benchmark with $390$ spoken prompts and $1{,}067$ human-authored binary audio and text rubrics spanning tone, persona, style/accent, and speed/length. On \textsc{SteerBench}, supervised training improves audio-steering average pass rate by $44.5$ percentage points over the strongest evaluated open baseline. On Audio MultiChallenge, task average pass rate improves by $7$ points over its strongest evaluated open baseline. RL further raises source-clean interruption response from $72.5\%$ to $82.5\%$ and reduces synthetic pause barge-in from $26.5\%$ to $9\%$. Steering and aggregate task scores remain comparable or higher, while reward probes reveal reward hacking through incomplete responses. Our model and benchmark support systematic research on spoken steerability, with reward analysis showing why timing gains must be evaluated alongside response completeness.
\end{abstract}
\endgroup

\section{Introduction}
\label{sec:intro}
Spoken dialogue conveys emotion, accent, and timing cues that text transcripts alone do not preserve. End-to-end models process this acoustic information directly~\citep{defossez2024moshi}, but useful conversational partners must also manage when to speak and follow instructions about content and vocal delivery.

\Needspace{9\baselineskip}
Full-duplex systems listen and speak simultaneously, enabling turn taking, backchanneling, and interruption handling within an ongoing conversation~\citep{wang2024fullduplex,hu2025salmduplex, defossez2024moshi}. Early systems established joint two-channel modeling~\citep{nguyen2023dgslm} while later work improved synchronization and stream interleaving~\citep{veluri2024syncllm,zhang2024omniflatten}. Fluent turn taking does not guarantee that systems follow user instructions about tone, persona, or delivery.

\begin{figure}[t]
\centering
\includegraphics[width=\linewidth]{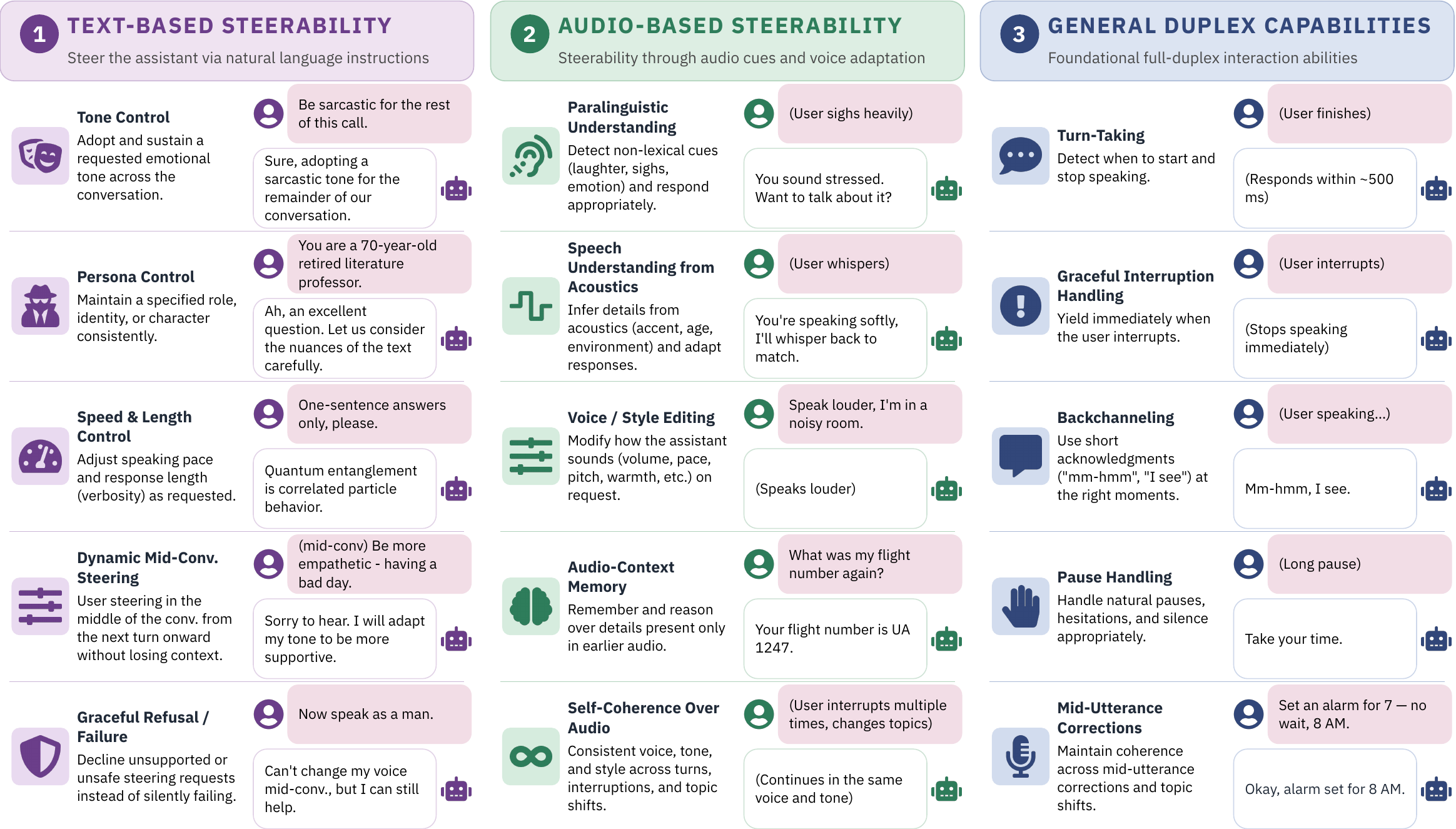}
\caption{\textbf{A capability taxonomy for full-duplex spoken dialogue.} The taxonomy distinguishes natural-language steering, acoustic understanding and adaptation, and general duplex interaction. Examples are schematic, not model outputs or latency measurements. The taxonomy describes a broader design space than the present benchmark: \textsc{SteerBench} tests requested content and delivery across tone, persona, style/accent, and speed/length; AudioMC, VoiceBench, and FDB provide complementary task and interaction evaluations.}
\label{fig:taxonomy}
\end{figure}

Hence, a useful interactive system must also be steerable along such attributes. Recent role- and voice-conditioned models make these attributes explicit control inputs~\citep{roy2026personaplex}. We use \emph{steerability} to mean the reliable shift of such behavior in response to user instructions. Text-model research distinguishes steerability from ordinary instruction following~\citep{chang2024steerability}, with controllable-generation methods ranging from explicit control codes~\citep{keskar2019ctrl} to broader adaptation approaches~\citep{liang2024ctgsurvey}. Evaluating this control in speech requires acoustic evidence, since speech-assessment studies distinguish lexical content from speech quality and paralinguistic features~\citep{manakul2025audiojudge}.

We organize spoken steerability into a capability taxonomy (Figure~\ref{fig:taxonomy}). Its three families connect \emph{what the model says}, \emph{how it sounds}, and \emph{how it participates}: natural-language steering, acoustic understanding and adaptation, and duplex interaction. The distinction matters because task compliance and floor management require separate checks: duplex benchmarks assess pauses, interruptions, and backchannels as distinct behaviors~\citep{lin2025fdbv1}. The taxonomy guides training and evaluation by assessing requested content and delivery alongside intelligibility, timing, and task capability.

To measure how steerable current full-duplex models are, we introduce \textsc{SteerBench}. Its $390$ spoken prompts combine concrete tasks with steering requests and use separate text and audio rubrics to assess content and delivery. Fixed reference clips anchor the requested acoustic style. Under matched items and the same judge, \textsc{Moshi} and \textsc{PersonaPlex} reach audio-steering average pass rates of only $20.55\%$ and $16.44\%$, respectively (Section~\ref{sec:steerability-bench}). These results motivate our central question: \textit{can targeted post-training make full-duplex models more steerable while preserving their general conversational capability?}

To study this question, we fine-tune \textsc{SteerDuplex}, built on the public \textsc{Moshi} backbone~\citep{defossez2024moshi}, on recorded conversations and synthetic speech/text examples covering instruction following, requested delivery, reasoning, safety, and duplex interaction (Section~\ref{sec:sft}). Continuing from this checkpoint, two-stage RL combines programmatic interaction rewards with judge-based transcript rubrics to optimize sampled speech continuations. Outcome-based and multi-signal rewards have proved useful in broader post-training~\citep{shao2024deepseekmath,lambert2024tulu3}; recent work extends them to speech timing and interaction~\citep{hsiao2026aspirin,ohashi2026interactivity}. Our design couples these objectives with incentives to sustain an answer until the user takes the floor.

Our reward design keeps content and delivery separate, since each requires different evidence. Text rubrics can judge whether an answer addresses a task, but cannot establish its prosody or speaking rate; audio-reference rubrics make delivery criteria concrete, and timing and waveform checks identify premature responses, silence, or invalid speech. The first RL stage pairs these timing and transcript rewards with response-continuity and waveform-validity checks; a second stage adds a continuation-duration bonus and dedicated user-backchannel sampling. This design is motivated by a recurring failure mode in our experiments, where a model improves interruption metrics by yielding too readily, including when it should continue speaking (Section~\ref{sec:analysis}).

\textsc{PersonaPlex} supports voice and role conditioning~\citep{roy2026personaplex}, while \textsc{F-Actor} learns instruction-controlled conversational behavior through supervised training~\citep{zufle2026factor}. ASPIRin isolates speaking decisions from token selection~\citep{hsiao2026aspirin}, and \citet{ohashi2026interactivity} combine interaction rewards with semantic feedback. Our contribution couples a steerability taxonomy with \textsc{SteerBench}, which checks requested content and reference-grounded vocal delivery separately. We combine targeted supervised steering with continuity-aware, two-stage RL and evaluate both alongside general task performance. Reward probes show how timing gains can mask incomplete responses.

Taken together, our contributions are fourfold:
\textbf{(i)} We propose \textsc{SteerDuplex}, a full-duplex speech model built around a taxonomy that connects instruction-based steering, acoustic delivery, and conversational interaction (Figure~\ref{fig:taxonomy}).
\textbf{(ii)} We introduce \textsc{SteerBench}, with $390$ spoken prompts and $1{,}067$ human-authored rubrics that assess content and reference-grounded delivery across tone, persona, style/accent, and speed/length (Section~\ref{sec:steerability-bench}).
\textbf{(iii)} We combine supervised fine-tuning, which establishes steering and task capability, with continuity-aware RL, which improves interruption response and pause handling while maintaining comparable or higher steering and aggregate task scores (Section~\ref{sec:main-results}).
\textbf{(iv)} We characterize reward hacking: isolated rewards admit empty speech, while interruption optimization can weaken continuation after listener feedback (Section~\ref{sec:analysis}).

\section{Related Work}
\label{sec:related}
\paragraph{Full-duplex speech models.}
Early two-channel dialogue modeling~\citep{nguyen2023dgslm} and subsequent synchronization and stream-interleaving methods~\citep{veluri2024syncllm,zhang2024omniflatten} established the basis for simultaneous listening and speaking. \textsc{Moshi} combines parallel user and assistant audio streams with a time-aligned text channel~\citep{defossez2024moshi}; later systems extend real-time duplex interaction~\citep{wang2024fullduplex,hu2025salmduplex}. \textsc{PersonaPlex} introduces role and voice conditioning~\citep{roy2026personaplex}, and \textsc{F-Actor} studies interactional control through supervised imitation~\citep{zufle2026factor}. We build on this model family to study requested content and delivery alongside conversational timing.

\paragraph{Steering and audio understanding.}
Controllable text generation conditions outputs on style, sentiment, or persona~\citep{keskar2019ctrl,dathathri2020pplm,liang2024ctgsurvey}. Steerability studies distinguish reliable behavioral control from ordinary instruction following~\citep{chang2024steerability}, using supervised adaptation~\citep{chen2024personality} or consistency rewards~\citep{abdulhai2025personas}. Extending this control to speech requires grounding instructions in acoustic information. Audio-language models have developed broader understanding and reasoning capabilities~\citep{ghosh2024gama}, while compositional and multi-task benchmarks test whether they can reason about relations and events in audio~\citep{ghosh2024compa,sakshi2025mmau}. These abilities help models interpret acoustic context; controlling their own delivery remains a separate problem.

\paragraph{Evaluating spoken interaction.}
Voice-assistant evaluations cover spoken instruction following~\citep{chen2024voicebench}, the integration of paralinguistic and visual cues~\citep{selvakumar2025multivox}, and retention of instructions and revisions across turns~\citep{gosai2025audiomc}. Adversarial audio-grounding tests additionally show that standard task scores can conceal responses unsupported by the input~\citep{seth2026aha}. Duplex benchmarks complement these evaluations with event timing and multi-turn task completion~\citep{lin2025fdbv1,lin2026fdbv2}. \textsc{SteerBench} adds content and delivery rubrics for explicit steering requests, alongside measures of task success and interaction.

\paragraph{Reinforcement learning for speech.}
GRPO and component-normalized variants optimize sampled outputs with multiple rewards~\citep{shao2024deepseekmath,ichihara2025mogrpo,liu2026gdpo}. Speech applications include alignment from annotated conversations~\citep{wu2025aligning,arora2026rlaif}, timing optimization~\citep{hsiao2026aspirin}, and joint alignment of duplex interaction behaviors~\citep{ohashi2026interactivity}. Reinforcement learning with verifiable rewards (RLVR) uses answer and constraint checks~\citep{lambert2024tulu3}; qualitative criteria require other forms of feedback. Policy-aware rubric weighting~\citep{tyagi2026pow3r} and rubric-conditioned self-distillation without a training-time verifier~\citep{rezaei2026rgsd} study these signals in general model post-training. For speech RL, the corresponding design problem also includes acoustic validity and timing: semantic feedback must reward responsiveness without encouraging incomplete answers.

\section{SteerDuplex: Post-Training for Speech Control}
\label{sec:methodology}
\label{sec:preliminaries}
\begin{figure}[t]
\centering
\includegraphics[width=\linewidth]{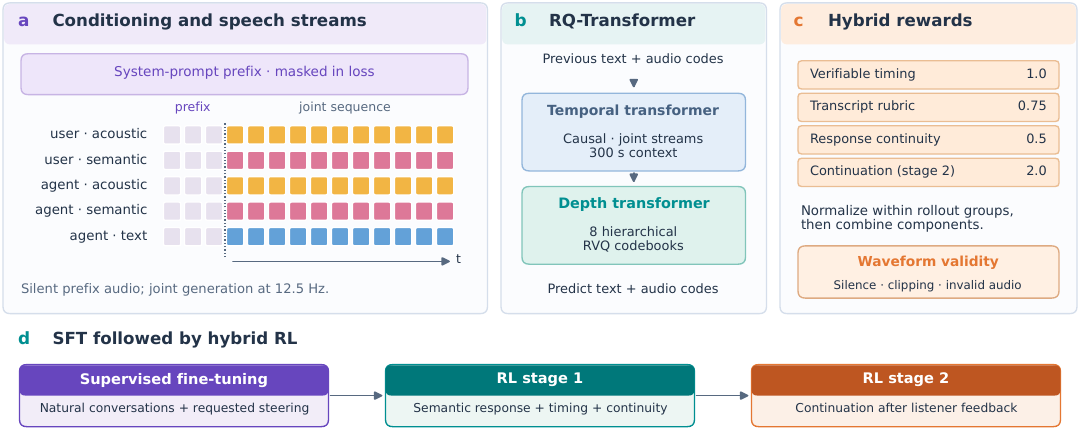}
\caption{\textbf{SteerDuplex architecture and training.} (a) A masked system-prompt prefix conditions joint audio/text streams. (b) Temporal and depth transformers predict hierarchical audio codes. (c) Grouped continuations receive component-normalized rewards and waveform validity checks. (d) SFT initializes RL stages rewarding response continuity, then continuation after listener feedback. Each RL stage freezes its initial model as the KL reference.}
\label{fig:method}
\end{figure}

\textsc{SteerDuplex} builds on the \textsc{Moshi} architecture~\citep{defossez2024moshi}. Supervised fine-tuning produces the model, \textsc{SteerDuplex}-SFT. Two-stage RL with hybrid rewards then refines its interaction behavior (Figure~\ref{fig:method}). We call the resulting checkpoint \textsc{SteerDuplex}-RL (``+ RL'' in tables); \textsc{SteerDuplex} alone denotes the supervised model. The hybrid objective combines verifiable programmatic interaction rewards (RLVR-style) with judge-based semantic rewards.

\subsection{Supervised Fine-Tuning}
\label{sec:architecture}
\label{sec:sft}
The model retains Moshi's temporal transformer, depth transformer, and hierarchical audio codec. A natural-language system prompt prefixes the assistant-text stream; both audio streams remain silent during the prefix, and a delimiter marks its end. We mask the prompt tokens in the training loss and use the same conditioning format throughout training and inference. We split conversations at alignment boundaries and cap context and response at $300$ seconds, subject to shorter benchmark limits.

The supervised training mixture combines natural conversations with targeted examples of instruction following, steering, duplex interaction, safety, and reasoning. Natural speech supplies variation in prosody, overlap, and turn structure; targeted examples supply explicit steering instructions and complete conversation state. The mixture contains $504{,}416$ audio records and $65{,}675$ text records, sampled at $88\%$ and $12\%$ of the training mass, respectively. Audio durations sum to $8{,}510.9$ hours of audio records; repeated source material means this is not a count of unique recording hours. Fixed sampling weights keep smaller targeted subsets represented without allowing synthetic examples to dominate. Quality checks cover semantic consistency, audio validity, provenance, privacy, and safety. We use word-level alignment from \texttt{Qwen3-ForcedAligner-0.6B}~\citep{shi2026qwen3asr} and turn-aware loss masks. Appendix~\ref{app:data-stats} gives data composition and Appendix~\ref{app:hparams} gives training settings.

\subsection{Two-Stage Reinforcement Learning}
\label{sec:rl-data}
\label{sec:opt}
RL optimizes sampled continuations from interaction windows spanning turns, interruptions, pauses, backchannels, noise, and speech-mirror scenarios. We filter CANDOR material to remove identified overlap with evaluation conversations. Each window contains aligned user audio, assistant history, transcripts, timing targets, and semantic grading metadata. Each rollout group shares one context.

For speech-text rollouts sharing a context, Group reward-Decoupled Normalization Policy Optimization (GDPO)~\citep{liu2026gdpo} normalizes each reward separately before combining components:
\begin{equation}
\hat{A}_i=\sum_k w_k\frac{R_i^{(k)}-\mathrm{mean}(\mathbf{R}^{(k)})}{\mathrm{std}(\mathbf{R}^{(k)})+\epsilon}.
\label{eq:gdpo-adv}
\end{equation}
Here $R_i^{(k)}$ is reward component $k$ for rollout $i$. This normalization prevents a component's raw scale from dominating the update. A component that is constant within a group contributes no learning signal.

\paragraph{Stage 1: response continuity.}
The first stage starts from the supervised checkpoint and uses it as a frozen KL reference. In addition to interaction timing and transcript rewards, it includes a response-continuity term with weight $0.5$ and a $4$-second first-response target. This term discourages short responses that satisfy a timing event but fail to sustain an answer.

\paragraph{Stage 2: continuation after listener feedback.}
The second stage initializes both policy and KL reference from the first-stage checkpoint. It retains the continuity term and adds a continuation-duration bonus of weight $2.0$, with a $4$-second target, on noise and user-backchannel events. Dedicated user-backchannel sampling emphasizes these events. This stage rewards continuing through listener feedback that does not request the floor. Both stages use a policy loss over text-stream actions and an adaptive sampled-action KL penalty, rather than the exact full-distribution KL of \citet{ohashi2026interactivity}. Moshi's temporal transformer processes joint audio/text history and supplies both text logits and the representation used by the audio decoder, so text-action gradients can change subsequent speech generation. We include sampled padding tokens because they encode frames without a new text token, making their timing relevant to pauses and speech onset; excluding them would remove credit from these decisions. Audio-codebook actions receive no direct policy loss. Appendix~\ref{app:hparams} gives the training settings.

\subsection{Rewards and Audio Validity}
\label{sec:rewards-design}
The RL objective combines interaction timing rewards (weight $1.0$), the continuity terms above, a Gemini 3.6 Flash transcript judge on turn and interruption groups (weight $0.75$), and a waveform-integrity gate. Timing rewards distinguish responding, yielding, waiting, and continuing; the judge assesses semantic response quality. Waveform checks reject silence, clipping, and invalid outputs. Interruption credit requires assistant speech before the interruption, preventing silence from earning yielding credit. Appendix~\ref{app:reward-weights} lists the weights; Section~\ref{sec:analysis} examines reward shortcuts and trade-offs.

\label{sec:ref-rewards}
Reference-audio rubrics evaluate steerability but do not enter the reported RL objective. Given a target clip, generated speech, and both transcripts, the judge assesses requested delivery, prosody, rate, articulation, naturalness, and task success. Speaker identity matters only when explicitly requested by the rubric. Fixed references give acoustic criteria a concrete target, although audio presentation and prompting can influence judge reliability~\citep{manakul2025audiojudge}.

\section{SteerBench: Evaluating Spoken Steerability}
\label{sec:steerability-bench}
\textsc{SteerBench} tests whether a model can satisfy a spoken steering request while completing a concrete task. It contains $390$ prompts across tone, persona, style/accent, and speed/length, with $1{,}067$ human-authored binary rubrics: $438$ audio rubrics and $629$ text rubrics. The test set is disjoint from the supervised and RL training data. Inference is audio-in/audio-out under the shared system prompt in Appendix~\ref{app:system-prompt}. Text rubrics are judged from transcripts; audio rubrics use fixed synthetic and human-sourced reference clips. Human reviewers validate references against the requested delivery. Appendices~\ref{app:audio-rubric-prompt} and~\ref{app:human-agreement} detail judging and human agreement.

We distinguish three aggregation levels. \emph{Audio-steering average pass rate (APR)} is the percentage of examples that satisfy every applicable audio rubric. \emph{Sample APR} additionally requires every text rubric to pass. \emph{Rubric pass rate} averages the individual binary decisions. These measures distinguish delivery control from full task compliance; passing many rubrics need not mean completing the task.\footnote{Passing three of four rubrics gives a $75\%$ rubric pass rate, but the example fails an all-constraints pass criterion.}

\begin{figure}[t]
\centering
\includegraphics[width=\linewidth]{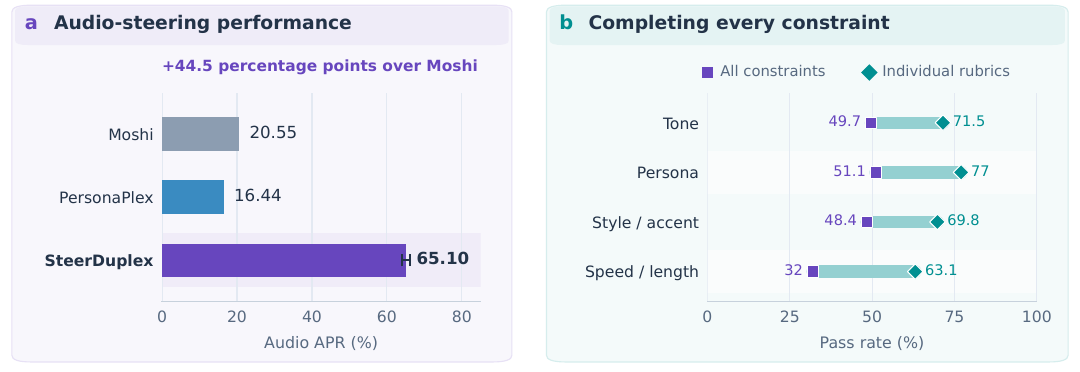}
\caption{\textbf{Steering improves, but complete task compliance remains difficult.} (a) Audio-steering APR on matched items. SFT reaches $65.10\pm1.13\%$ over three runs; baseline scores retain their source protocols. (b) These same three runs supply sample APR and individual-rubric pass rates. Gemini 3.6 Flash judges both panels; see Table~\ref{tab:steerability}.}
\label{fig:steerbench}
\end{figure}

Under matched items, rubric implementation, and the Gemini 3.6 Flash judge, the supervised model reaches $65.10\pm1.13\%$ audio-steering APR, compared with $20.55\%$ for \textsc{Moshi} and $16.44\%$ for \textsc{PersonaPlex} (Figure~\ref{fig:steerbench}). Sample APR ranges from $32\%$ to $51.11\%$; individual-rubric pass rates range from $63.10\%$ to $77.03\%$. Reliably satisfying every constraint remains difficult.

\section{Evaluation Setup}
\label{sec:experiments}
\label{sec:setup}
\paragraph{Benchmarks and baselines.}
Alongside \textsc{SteerBench}, we evaluate multi-turn robustness with Audio MultiChallenge~\citep{gosai2025audiomc}, spoken instruction following with VoiceBench~\citep{chen2024voicebench}, and interaction with Full-Duplex-Bench (FDB)~\citep{lin2025fdbv1,lin2026fdbv2}. FDB-v1 tests individual events; its v1.5 extension tests interruption and overlap handling~\citep{lin2025fdbv15}; FDB-v2 evaluates multi-turn task sessions. We compare with \textsc{Moshi} and \textsc{PersonaPlex} to assess steerability gains alongside maintained or improved task performance and duplex interaction. Published hosted-model results provide context, with model identities and protocol differences stated in the tables. Repeated benchmark evaluations report means and population standard deviations across three decoding runs. Published baseline scores retain their source protocols. See Appendix~\ref{app:baselines}.

\paragraph{Source overlap.}
Some official FDB-v1 turn-taking and pause examples reuse CANDOR conversations present in supervised training. We treat those scores as diagnostics. To assess generalization, we use FDB-v2, the $498$-example FDB-v1.5 split without Fisher/CANDOR material, and FDB-v1's source-clean synthetic interruption, synthetic pause, and backchannel tasks.

\paragraph{Checkpoint selection.}
We select the RL checkpoint on a separate development set targeting the interaction and task capabilities measured by the benchmarks. We freeze the checkpoint before benchmark evaluation; benchmark test scores do not influence checkpoint choice (Appendix~\ref{app:protocol}).
\section{Results}
\label{sec:main-results}
The supervised model establishes steering and task capability. RL then improves interruption response and pause handling with comparable or higher steering and aggregate task means. We compare task performance before examining these gains and costs.

\subsection{Multi-Turn and General Spoken Capability}
\begin{table}[t]
\centering
\caption{\textbf{Supervised task performance.} SteerDuplex averages three runs per benchmark; baseline scores retain their source protocols; $\pm$ denotes population standard deviation. AudioMC scores are percentages; FDB-v2 uses a $1$--$5$ scale; VoiceBench retains its official scales and paired subscores. Higher is better; bold marks the best displayed score per column, with paired subscores compared separately. AudioMC axes: instruction memory (IM), instruction revision (IR), self-correction (SC), and voice editing (VE).}
\label{tab:capabilities}
\footnotesize
\setlength{\tabcolsep}{3.2pt}
\noindent\textbf{(a) Audio MultiChallenge}\par\vspace{5pt}
\begin{tabularx}{\linewidth}{@{}l*{6}{>{\raggedleft\arraybackslash}X}@{}}
\toprule
Model & APR (\%) & ARS (\%) & IM & IR & SC & VE \\
\midrule
\textsc{Moshi} & 3.98 & 14.29 & 9.09 & 2.50 & 3.61 & 0 \\
\textsc{PersonaPlex} & 6.64 & 20.56 & 10.61 & 6.67 & 9.64 & 0 \\
\rowcolor{scaleBlue!8}
\textsc{SteerDuplex} & \textbf{13.64$\pm$0.28} & \textbf{37.17$\pm$1.33} & \textbf{17.93} & \textbf{14.17} & \textbf{21.69} & \textbf{2.56} \\
\bottomrule
\end{tabularx}
\par\vspace{9pt}
\noindent\textbf{(b) Full-Duplex-Bench v2}\par\vspace{5pt}
\begin{tabularx}{\linewidth}{@{}l*{5}{>{\raggedleft\arraybackslash}X}@{}}
\toprule
Model & Correction & Daily & Entity & Safety & Mean \\
\midrule
\textsc{Moshi} & 2.78 & 2.26 & 2.63 & 2.68 & 2.59 \\
\textsc{PersonaPlex} & 3.38 & 2.06 & 2.51 & 2.66 & 2.65 \\
\rowcolor{scaleBlue!8}
\textsc{SteerDuplex} & \textbf{4.21} & \textbf{3.74} & \textbf{4.07} & \textbf{4.65} & \textbf{4.17} \\
\bottomrule
\end{tabularx}
\par\vspace{9pt}
\noindent\textbf{(c) VoiceBench}\par\vspace{5pt}
\begin{tabularx}{\linewidth}{@{}l>{\hsize=0.8\hsize\linewidth=\hsize\centering\arraybackslash}X>{\hsize=0.8\hsize\linewidth=\hsize\centering\arraybackslash}X>{\hsize=0.9\hsize\linewidth=\hsize\centering\arraybackslash}X>{\hsize=1.5\hsize\linewidth=\hsize\centering\arraybackslash}X>{\hsize=1.4\hsize\linewidth=\hsize\centering\arraybackslash}X>{\hsize=0.75\hsize\linewidth=\hsize\centering\arraybackslash}X>{\hsize=0.85\hsize\linewidth=\hsize\centering\arraybackslash}X>{\hsize=1.0\hsize\linewidth=\hsize\centering\arraybackslash}X@{}}
\toprule
Model & Alpaca & Common & WildVoice & SD-QA & IFEval & BBH & AdvBench & Overall \\
\midrule
\textsc{Moshi} & 2.15 & 1.94 & 1.49 & 25.4 / \textbf{35.3} & 9.9 / 19.4 & \textbf{49.9} & 68.2 & 38.55 \\
\textsc{PersonaPlex} & \textbf{2.45} & \textbf{2.29} & 1.56 & 20.6 / 27.8 & 9 / 17.2 & 48.6 & 5.3 & 30.51 \\
\rowcolor{scaleBlue!8}
\textsc{SteerDuplex} & 2.28 & 2.21 & \textbf{1.74} & \textbf{26.88} / 26.88 & \textbf{10} / \textbf{19.7} & \textbf{49.9} & \textbf{99.3} & \textbf{40.87$\pm$0.27} \\
\bottomrule
\end{tabularx}
\par\vspace{9pt}
\end{table}

On Audio MultiChallenge, \textsc{SteerDuplex} obtains $13.64\pm0.28\%$ APR and $37.17\pm1.33\%$ average rubric score (ARS), compared with $6.64\%$ and $20.56\%$ for the strongest open baseline, \textsc{PersonaPlex} (Table~\ref{tab:capabilities}). Fully successful multi-turn tasks remain uncommon, particularly those requiring spoken revisions.

On FDB-v2, the supervised model scores $4.17$ under the slow examiner, exceeding \textsc{Moshi} ($2.59$) and \textsc{PersonaPlex} ($2.65$) in every task family. Its VoiceBench mean is $40.87\pm0.27$, compared with $38.55$ for \textsc{Moshi} and $30.51$ for \textsc{PersonaPlex}. Table~\ref{tab:capabilities} gives the task breakdowns; the same SFT runs serve as the reference for RL within each benchmark.

\subsection{RL Improves Interruption Response and Pause Handling}
\begin{figure}[t]
\centering
\includegraphics[width=\linewidth]{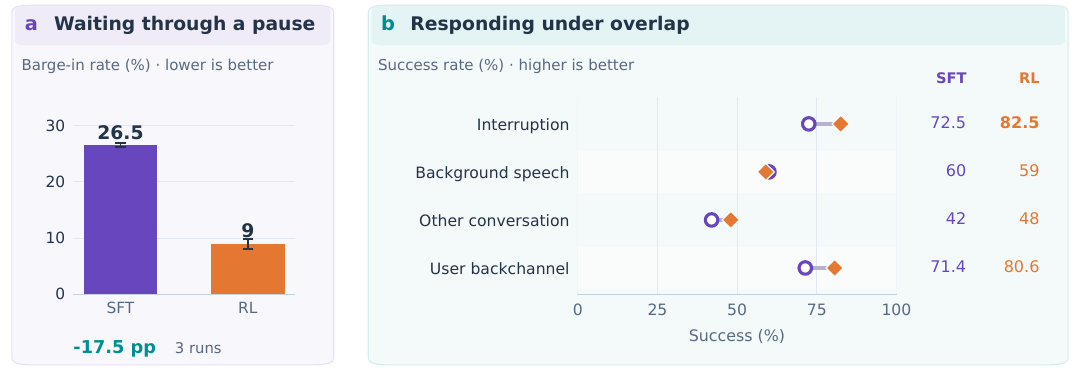}
\caption{\textbf{RL improves when to wait and when to respond.} (a) Source-clean synthetic pause barge-in averaged across three runs per model; error bars show population standard deviation. (b) FDB-v1.5 success on $498$ paired examples. Table~\ref{tab:fdb_v15} reports the condition sizes and success rates.}
\label{fig:overlap}
\end{figure}

On source-clean FDB-v1.5, RL raises correct response after interruption from $72.5\%$ to $82.5\%$ (Figure~\ref{fig:overlap}). Continuation after a user backchannel rises from $71.4\%$ to $80.6\%$. Background-speech recovery changes from $60\%$ to $59\%$; recovery after speech directed elsewhere rises from $42\%$ to $48\%$.

On synthetic pause items, barge-in falls from $26.5\%$ to $9\%$, a difference of $-17.5$ percentage points. On the synthetic interruption task, response rate rises from $96\%$ to $97.7\%$, while semantic score changes from $3.94$ to $3.88$ and mean takeover latency increases by $40$\,ms (Table~\ref{tab:retention-details}). Thus better response timing does not imply uniform improvement in every interaction measure.

\subsection{Steering, Task Scores, and Interaction Costs}
\begin{figure}[t]
\centering
\includegraphics[width=\linewidth]{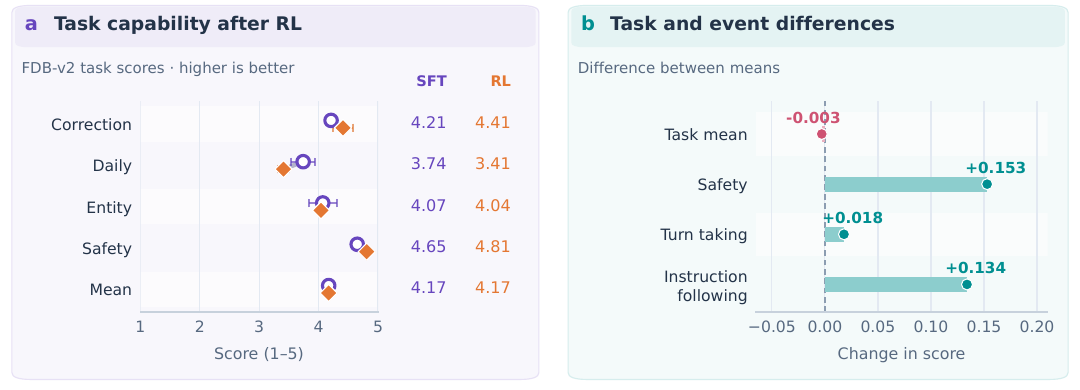}
\caption{\textbf{Task capability.} (a) FDB-v2 task-family means; error bars show population standard deviation across three runs per model. (b) Differences between means for task and event scores. Each model uses the same three runs in both panels. Colors indicate the direction of the point difference, not statistical significance. Higher is better. Table~\ref{tab:retention-details} reports dispersion.}
\label{fig:retention}
\end{figure}

\begin{table}[t]
\centering
\begin{minipage}{0.70\textwidth}
\centering
\captionsetup{justification=raggedright,singlelinecheck=false}
\caption{\textbf{Capability and interaction.} Higher is better; bold marks best displayed scores, including ties. Three runs per model; full dispersion is in Table~\ref{tab:retention-details}.}
\label{tab:posttraining-retention}
\footnotesize
\setlength{\tabcolsep}{5pt}
\begin{tabularx}{\linewidth}{@{}Xrr@{}}
\toprule
Metric & SFT & + RL \\
\midrule
\multicolumn{3}{@{}l}{\textit{Steering and task capability}} \\
\textsc{SteerBench} rubric pass (\%) & 63.75 & \textbf{65.22} \\
AudioMC APR (\%) & 13.64 & \textbf{14.38} \\
VoiceBench overall & 40.87 & \textbf{41.38} \\
FDB-v2 task mean & \textbf{4.17} & \textbf{4.17} \\
\midrule
\multicolumn{3}{@{}l}{\textit{Conversational behavior}} \\
FDB-v2 turn taking & 4.18 & \textbf{4.19} \\
FDB-v2 instruction following & 3.67 & \textbf{3.81} \\
\bottomrule
\end{tabularx}

\end{minipage}
\end{table}
The comparison (Table~\ref{tab:posttraining-retention}) gives comparable or higher steering and aggregate task means. VoiceBench rises from $40.87$ to $41.38$, and FDB-v2 safety from $4.65$ to $4.81$. FDB-v2 mean task score changes by only $-0.003$; its per-event turn-taking and instruction-following differences are $+0.018$ and $+0.134$ (Figure~\ref{fig:retention}). These task-level means do not establish that every conversational behavior is retained.

RL also takes over the floor less often during user backchannels ($4.8\%$ versus $9.1\%$ for SFT). However, aggregate task scores can conceal premature yielding in other contexts. Section~\ref{sec:analysis} examines this tension between holding and yielding the floor.

\section{Reward Design and Optimization Dynamics}
\label{sec:analysis}
\label{sec:rl-arms}
We examine response continuity, reward hacking, and sensitivity to training choices. Appendix~\ref{app:ablations} gives the full comparisons and judge checks.

\subsection{Continuity Supports Sustained Responses}
\label{sec:rl-arm-table}
Without our continuity additions, an SFT-initialized control using the original interaction recipe continues for only $1.10$ seconds after a user backchannel in the fixed development diagnostic. FDB-v2 turn taking scores $3.93$, versus $4.18$ for SFT, $4.04$ after stage 1, and $4.19$ after stage 2. Differences in interaction pools, optimization settings, and training budgets preclude attributing these gains solely to continuity rewards.

The second stage sustains longer responses: SFT, stage 1, and RL average $22.1$, $20.1$, and $23.0$ words per utterance across three FDB-v2 daily-task runs. Development continuation after a user backchannel rises from SFT's $2.60$ seconds to $3.20$ seconds. Yet RL ends daily-task utterances while the examiner is speaking more often than SFT ($32.3\%$ versus $23.6\%$), despite longer responses.

\subsection{Reward Hacking in Duplex Speech}
\label{sec:hacking}
A model that remains silent can earn interruption credit without ever yielding. Requiring speech before the interruption closes this shortcut; continuity rewards additionally encourage sustained speech when the user has not taken the floor.

\begin{figure}[t]
\centering
\includegraphics[width=\linewidth]{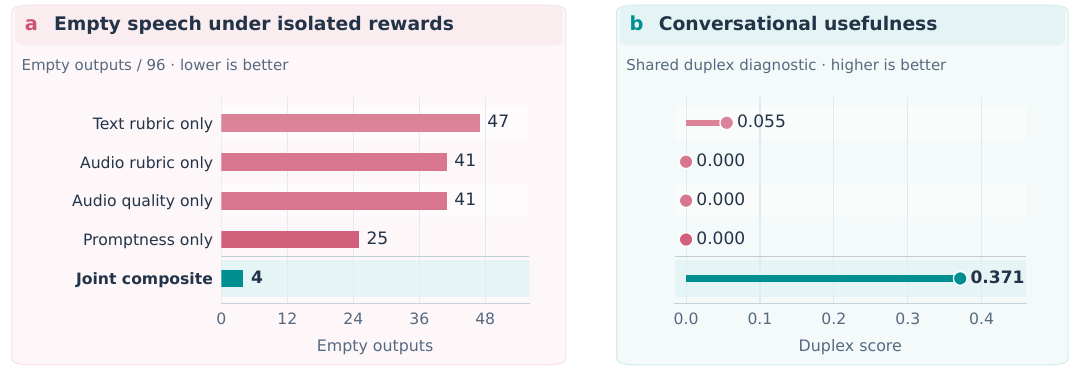}
\caption{\textbf{Isolated rewards admit empty or unresponsive speech.} Separate reward probes each evaluate $96$ held-out sampled rollouts. Left: empty outputs; lower is better. Right: the shared duplex diagnostic; higher is better. The joint composite improves both diagnostics but still leaves four empty outputs. These probes are separate from the reported RL model. Scalar training rewards use different objectives and should not be compared across conditions. Table~\ref{tab:probe-summary} gives the full results.}
\label{fig:reward-probes}
\end{figure}

In separate reward probes (Figure~\ref{fig:reward-probes}), promptness-only optimization attains a scalar reward of $0.670$ but scores zero on the shared duplex diagnostic, leaving $25$ of $96$ rollouts empty. Text-only, audio-rubric-only, and audio-quality-only objectives leave $47$, $41$, and $41$ empty rollouts. The joint composite improves the duplex score to $0.371$ but still leaves $4$ empty outputs. Appendix~\ref{app:reward-hacking-details} details these probes, which are separate from the reported interaction RL runs.

\subsection{Continued Optimization Can Erode Response Continuity}
\label{sec:rl-dose}
Further optimization can improve yielding while weakening continuation through listener feedback.

\begin{figure}[t]
\centering
\includegraphics[width=\linewidth]{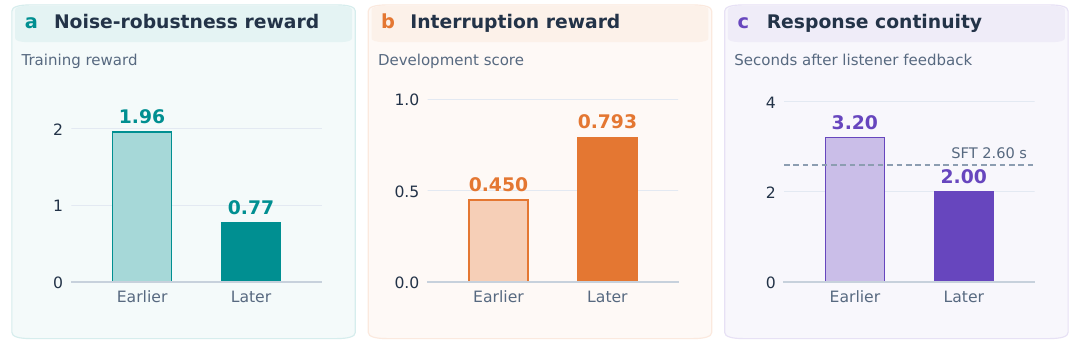}
\caption{\textbf{Higher interruption reward can coexist with weaker response continuity.} Recorded snapshots from the second RL stage. (a) The noise-robustness reward, which includes continuation, declines with further optimization. (b) The interruption reward on natural development conversations rises, while (c) continuation after a user backchannel falls below the SFT reference.}
\label{fig:reward-integrity}
\end{figure}

In the second stage (Figure~\ref{fig:reward-integrity}), interruption reward on natural development conversations rises from $0.450$ to $0.793$, while continuation after a user backchannel falls from $3.20$ to $2.00$ seconds and noise-robustness reward declines from $1.96$ to $0.77$. These measurements suggest interference between interaction objectives without isolating its mechanism. Group normalization balances scales, but any component that becomes constant within a rollout group supplies no gradient to resist other reward components.

\subsection{Robustness and Behavioral Trade-offs}
A training-seed replicate reaches $9\%$ synthetic pause barge-in and $4.22$ FDB-v2 turn taking. Applying the final reward directly from SFT yields $16.8\%$ and $4.13$ with less training, so staging is not isolated.

\label{sec:judge-sensitivity}
Rescoring identical generations from three runs per model gives RL-minus-SFT turn-taking differences of $-0.074$ under Gemini 3.6 Flash and $-0.059$ under gpt-5.4-mini at low reasoning effort (Table~\ref{tab:crossjudge-fdb2}). Both judges find a loss in this common pool, which differs from the broader comparison in Table~\ref{tab:retention-details}. \textsc{SteerBench} and AudioMC use one judge.

\section{Limitations}
\label{sec:limitations}
\paragraph{Scope and selection.}
\textsc{SteerBench} covers controlled English requests with fixed reference clips, rather than unrestricted personalization across languages, dialects, and recording conditions. Some fine-grained styles remain below the stronger open baseline. Repeated evaluations average the three highest-scoring runs from pools of unequal size, potentially inflating scores; FDB-v1.5 uses one decoding pass per checkpoint. CANDOR overlap further limits the use of official FDB-v1 turn and pause results.

\paragraph{Interaction trade-offs.}
The reported RL checkpoint yields too readily in some FDB-v2 contexts. The single-stage comparison has fewer updates than the staged run, so it cannot establish that staging itself causes the observed advantage. The first-stage FDB-v1.5 row omits an unavailable acoustic check; latency distributions beyond means are unavailable.

\paragraph{Backbone capacity.}
SteerDuplex retains Moshi's $7$B model, audio codec, and streaming architecture~\citep{defossez2024moshi}. These components constrain the reasoning, acoustic representation, and latency available to post-training. Our results measure improvement within that backbone, rather than the ceiling of steerable speech models. We have not isolated the effects of the backbone, training data, and reward design.

\paragraph{Judges and deployment.}
Rubric rewards are imperfect proxies: they depend on the LLM judge's ability to recognize the requested characteristics of a correct response. Reference-audio evaluation also depends on the judge's comparison with a target clip, which can favor that particular realization of a style even after human validation. Judge errors can therefore affect both the learning signal and reported scores. The second-judge study agrees on a turn-taking loss within its common run pool, while the broader comparison gives a small positive difference; no such study is available for \textsc{SteerBench} or AudioMC. Automated safety and audio checks also do not establish deployment safety. Appendix~\ref{app:data-protocol} covers consent, data handling, and misuse.

\section{Conclusion}
\label{sec:conclusion}
Our steerability taxonomy organizes content, delivery, and interaction control. \textsc{SteerDuplex} improves these capabilities, and \textsc{SteerBench} evaluates requested content and delivery separately. Supervised fine-tuning supplies the main steerability gains; RL refines interaction behavior. Explicit continuity rewards reduce the tendency to obtain timing credit through short or absent responses, but do not eliminate the conflict between yielding and continuing. Premature yielding calls for evaluation of conversational events alongside aggregate task success.

\renewcommand{\bibfont}{\footnotesize}
\setlength{\bibsep}{1pt}
\bibliographystyle{abbrvnat}
\bibliography{custom}

\begin{thebibliography}{40}
\providecommand{\natexlab}[1]{#1}
\providecommand{\url}[1]{\texttt{#1}}
\expandafter\ifx\csname urlstyle\endcsname\relax
  \providecommand{\doi}[1]{doi: #1}\else
  \providecommand{\doi}{doi: \begingroup \urlstyle{rm}\Url}\fi

\bibitem[Abdulhai et~al.(2025)Abdulhai, Cheng, Clay, Althoff, Levine, and
  Jaques]{abdulhai2025personas}
M.~Abdulhai, R.~Cheng, D.~Clay, T.~Althoff, S.~Levine, and N.~Jaques.
\newblock Consistently simulating human personas with multi-turn reinforcement
  learning.
\newblock In \emph{Advances in Neural Information Processing Systems}, 2025.
\newblock URL \url{https://arxiv.org/abs/2511.00222}.

\bibitem[Arora et~al.(2026)Arora, Tian, Shi, Futami, Kashiwagi, Tsunoo, and
  Watanabe]{arora2026rlaif}
S.~Arora, J.~Tian, J.~Shi, H.~Futami, Y.~Kashiwagi, E.~Tsunoo, and S.~Watanabe.
\newblock Optimizing conversational quality in spoken dialogue systems with
  reinforcement learning from {AI} feedback.
\newblock \emph{arXiv preprint arXiv:2601.19063}, 2026.
\newblock URL \url{https://arxiv.org/abs/2601.19063}.

\bibitem[Chang et~al.(2024)Chang, Wiens, Schnabel, and
  Swaminathan]{chang2024steerability}
T.~Chang, J.~Wiens, T.~Schnabel, and A.~Swaminathan.
\newblock Measuring steerability in large language models.
\newblock In \emph{NeurIPS Workshop on Safe Generative {AI}}, 2024.
\newblock URL \url{https://openreview.net/forum?id=y2J5dAqcJW}.

\bibitem[Chen et~al.(2024{\natexlab{a}})Chen, Wu, Guo, Huang, and
  Dai]{chen2024personality}
Y.~Chen, Z.~Wu, J.~Guo, S.~Huang, and X.~Dai.
\newblock Extroversion or introversion? {C}ontrolling the personality of your
  large language models.
\newblock \emph{arXiv preprint arXiv:2406.04583}, 2024{\natexlab{a}}.
\newblock URL \url{https://arxiv.org/abs/2406.04583}.

\bibitem[Chen et~al.(2024{\natexlab{b}})Chen, Yue, Zhang, Gao, Tan, and
  Li]{chen2024voicebench}
Y.~Chen, X.~Yue, C.~Zhang, X.~Gao, R.~T. Tan, and H.~Li.
\newblock {VoiceBench}: Benchmarking {LLM}-based voice assistants.
\newblock \emph{arXiv preprint arXiv:2410.17196}, 2024{\natexlab{b}}.
\newblock URL \url{https://arxiv.org/abs/2410.17196}.

\bibitem[Cieri et~al.(2004)Cieri, Miller, and Walker]{cieri2004fisher}
C.~Cieri, D.~Miller, and K.~Walker.
\newblock The {Fisher} corpus: A resource for the next generations of
  speech-to-text.
\newblock In \emph{Proceedings of the 4th International Conference on Language
  Resources and Evaluation (LREC)}, 2004.
\newblock URL \url{https://aclanthology.org/L04-1500/}.

\bibitem[Dathathri et~al.(2020)Dathathri, Madotto, Lan, Hung, Frank, Molino,
  Yosinski, and Liu]{dathathri2020pplm}
S.~Dathathri, A.~Madotto, J.~Lan, J.~Hung, E.~Frank, P.~Molino, J.~Yosinski,
  and R.~Liu.
\newblock Plug and play language models: A simple approach to controlled text
  generation.
\newblock In \emph{International Conference on Learning Representations}, 2020.
\newblock URL \url{https://arxiv.org/abs/1912.02164}.

\bibitem[D{\'e}fossez et~al.(2024)D{\'e}fossez, Mazar{\'e}, Orsini, Royer,
  P{\'e}rez, J{\'e}gou, Grave, and Zeghidour]{defossez2024moshi}
A.~D{\'e}fossez, L.~Mazar{\'e}, M.~Orsini, A.~Royer, P.~P{\'e}rez,
  H.~J{\'e}gou, E.~Grave, and N.~Zeghidour.
\newblock Moshi: A speech-text foundation model for real-time dialogue.
\newblock \emph{arXiv preprint arXiv:2410.00037}, 2024.
\newblock URL \url{https://arxiv.org/abs/2410.00037}.

\bibitem[Ghosh et~al.(2024{\natexlab{a}})Ghosh, Kumar, Seth, Evuru, Tyagi,
  Sakshi, Nieto, Duraiswami, and Manocha]{ghosh2024gama}
S.~Ghosh, S.~Kumar, A.~Seth, C.~K.~R. Evuru, U.~Tyagi, S.~Sakshi, O.~Nieto,
  R.~Duraiswami, and D.~Manocha.
\newblock {GAMA}: A large audio-language model with advanced audio
  understanding and complex reasoning abilities.
\newblock In \emph{Proceedings of the 2024 Conference on Empirical Methods in
  Natural Language Processing}, pages 6288--6313, Miami, Florida, USA,
  2024{\natexlab{a}}. Association for Computational Linguistics.
\newblock \doi{10.18653/v1/2024.emnlp-main.361}.
\newblock URL \url{https://aclanthology.org/2024.emnlp-main.361/}.

\bibitem[Ghosh et~al.(2024{\natexlab{b}})Ghosh, Seth, Kumar, Tyagi, Evuru,
  Ramaneswaran, Sakshi, Nieto, Duraiswami, and Manocha]{ghosh2024compa}
S.~Ghosh, A.~Seth, S.~Kumar, U.~Tyagi, C.~K. Evuru, S.~Ramaneswaran, S.~Sakshi,
  O.~Nieto, R.~Duraiswami, and D.~Manocha.
\newblock {CompA}: Addressing the gap in compositional reasoning in
  audio-language models.
\newblock In \emph{International Conference on Learning Representations},
  2024{\natexlab{b}}.
\newblock URL \url{https://arxiv.org/abs/2310.08753}.

\bibitem[{Google}(2026)]{google2026gemini36flash}
{Google}.
\newblock {Gemini 3.6 Flash}.
\newblock Google AI for Developers model documentation, 2026.
\newblock URL
  \url{https://ai.google.dev/gemini-api/docs/models/gemini-3.6-flash}.
\newblock Model ID \texttt{gemini-3.6-flash}; accessed 2026-09-01.

\bibitem[{Google DeepMind}(2025)]{google2025geminilive}
{Google DeepMind}.
\newblock {Gemini Live API} overview.
\newblock Google AI for Developers documentation, 2025.
\newblock URL \url{https://ai.google.dev/gemini-api/docs/live-api}.
\newblock Accessed 2026-09-09.

\bibitem[Gosai et~al.(2025)Gosai, Vuong, Tyagi, Li, You, Bavare, U{\c{c}}ar,
  Fang, Jang, Liu, and He]{gosai2025audiomc}
A.~Gosai, T.~Vuong, U.~Tyagi, S.~Li, W.~You, M.~Bavare, A.~U{\c{c}}ar, Z.~Fang,
  B.~Jang, B.~Liu, and Y.~He.
\newblock Audio {MultiChallenge}: A multi-turn evaluation of spoken dialogue
  systems on natural human interaction.
\newblock \emph{arXiv preprint arXiv:2512.14865}, 2025.
\newblock URL \url{https://arxiv.org/abs/2512.14865}.

\bibitem[Hsiao et~al.(2026)Hsiao, Lu, Fu, Lin, Hung, and Lee]{hsiao2026aspirin}
C.-Y. Hsiao, K.-H. Lu, Y.-K. Fu, G.-T. Lin, H.-T. Hung, and H.-y. Lee.
\newblock {ASPIRin}: Action space projection for interactivity-optimized
  reinforcement learning in full-duplex speech language models.
\newblock \emph{arXiv preprint arXiv:2604.10065}, 2026.
\newblock URL \url{https://arxiv.org/abs/2604.10065}.

\bibitem[Hu et~al.(2025)Hu, Hosseini-Asl, Chen, Casanova, Ghosh, {\.Z}elasko,
  Chen, Li, Balam, and Ginsburg]{hu2025salmduplex}
K.~Hu, E.~Hosseini-Asl, C.~Chen, E.~Casanova, S.~Ghosh, P.~{\.Z}elasko,
  Z.~Chen, J.~Li, J.~Balam, and B.~Ginsburg.
\newblock {SALM-Duplex}: Efficient and direct duplex modeling for
  speech-to-speech language model.
\newblock In \emph{Proc.\ Interspeech}, 2025.
\newblock URL \url{https://arxiv.org/abs/2505.15670}.

\bibitem[Ichihara et~al.(2025)Ichihara, Jinnai, Morimura, Sakamoto, Mitsuhashi,
  and Uchibe]{ichihara2025mogrpo}
Y.~Ichihara, Y.~Jinnai, T.~Morimura, M.~Sakamoto, R.~Mitsuhashi, and E.~Uchibe.
\newblock {MO-GRPO}: Mitigating reward hacking of group relative policy
  optimization on multi-objective problems.
\newblock \emph{arXiv preprint arXiv:2509.22047}, 2025.
\newblock URL \url{https://arxiv.org/abs/2509.22047}.

\bibitem[Keskar et~al.(2019)Keskar, McCann, Varshney, Xiong, and
  Socher]{keskar2019ctrl}
N.~S. Keskar, B.~McCann, L.~R. Varshney, C.~Xiong, and R.~Socher.
\newblock {CTRL}: A conditional transformer language model for controllable
  generation.
\newblock \emph{arXiv preprint arXiv:1909.05858}, 2019.
\newblock URL \url{https://arxiv.org/abs/1909.05858}.

\bibitem[Lambert et~al.(2024)Lambert, Morrison, Pyatkin, Huang, Ivison,
  Brahman, Miranda, Liu, Dziri, Lyu, et~al.]{lambert2024tulu3}
N.~Lambert, J.~Morrison, V.~Pyatkin, S.~Huang, H.~Ivison, F.~Brahman, L.~J.~V.
  Miranda, A.~Liu, N.~Dziri, S.~Lyu, et~al.
\newblock {T\"{u}lu~3}: Pushing frontiers in open language model post-training.
\newblock \emph{arXiv preprint arXiv:2411.15124}, 2024.
\newblock URL \url{https://arxiv.org/abs/2411.15124}.

\bibitem[Liang et~al.(2024)Liang, Wang, Wang, Song, Yang, Niu, Hu, Liu, Yao,
  Xiong, and Li]{liang2024ctgsurvey}
X.~Liang, H.~Wang, Y.~Wang, S.~Song, J.~Yang, S.~Niu, J.~Hu, D.~Liu, S.~Yao,
  F.~Xiong, and Z.~Li.
\newblock Controllable text generation for large language models: A survey.
\newblock \emph{arXiv preprint arXiv:2408.12599}, 2024.
\newblock URL \url{https://arxiv.org/abs/2408.12599}.

\bibitem[Lin et~al.(2025{\natexlab{a}})Lin, Kuan, Shi, Chang, Arora, Watanabe,
  and Lee]{lin2026fdbv2}
G.-T. Lin, S.-Y.~S. Kuan, J.~Shi, K.-W. Chang, S.~Arora, S.~Watanabe, and H.-y.
  Lee.
\newblock {Full-Duplex-Bench-v2}: A multi-turn evaluation framework for duplex
  dialogue systems with an automated examiner.
\newblock \emph{arXiv preprint arXiv:2510.07838}, 2025{\natexlab{a}}.
\newblock URL \url{https://arxiv.org/abs/2510.07838}.

\bibitem[Lin et~al.(2025{\natexlab{b}})Lin, Kuan, Wang, Lian, Li, Watanabe, and
  Lee]{lin2025fdbv15}
G.-T. Lin, S.-Y.~S. Kuan, Q.~Wang, J.~Lian, T.~Li, S.~Watanabe, and H.-y. Lee.
\newblock {Full-Duplex-Bench} v1.5: Evaluating overlap handling for full-duplex
  speech models.
\newblock \emph{arXiv preprint arXiv:2507.23159}, 2025{\natexlab{b}}.
\newblock URL \url{https://arxiv.org/abs/2507.23159}.

\bibitem[Lin et~al.(2025{\natexlab{c}})Lin, Lian, Li, Wang, Anumanchipalli,
  Liu, and Lee]{lin2025fdbv1}
G.-T. Lin, J.~Lian, T.~Li, Q.~Wang, G.~Anumanchipalli, A.~H. Liu, and H.-y.
  Lee.
\newblock {Full-Duplex-Bench}: A benchmark to evaluate full-duplex spoken
  dialogue models on turn-taking capabilities.
\newblock In \emph{Proc.\ ASRU}, 2025{\natexlab{c}}.
\newblock URL \url{https://arxiv.org/abs/2503.04721}.

\bibitem[Liu et~al.(2026)Liu, Dong, Lu, Diao, Belcak, Liu, Chen, Yin, Wang,
  Cheng, Choi, Kautz, and Molchanov]{liu2026gdpo}
S.-Y. Liu, X.~Dong, X.~Lu, S.~Diao, P.~Belcak, M.~Liu, M.-H. Chen, H.~Yin,
  Y.-C.~F. Wang, K.-T. Cheng, Y.~Choi, J.~Kautz, and P.~Molchanov.
\newblock {GDPO}: Group reward-decoupled normalization policy optimization for
  multi-reward {RL} optimization.
\newblock \emph{arXiv preprint arXiv:2601.05242}, 2026.
\newblock URL \url{https://arxiv.org/abs/2601.05242}.

\bibitem[Manakul et~al.(2025)Manakul, Gan, Ryan, Khan, Sirichotedumrong,
  Pipatanakul, Held, and Yang]{manakul2025audiojudge}
P.~Manakul, W.~H. Gan, M.~J. Ryan, A.~S. Khan, W.~Sirichotedumrong,
  K.~Pipatanakul, W.~Held, and D.~Yang.
\newblock {AudioJudge}: Understanding what works in large audio model based
  speech evaluation.
\newblock \emph{arXiv preprint arXiv:2507.12705}, 2025.
\newblock URL \url{https://arxiv.org/abs/2507.12705}.

\bibitem[Nguyen et~al.(2023)Nguyen, Kharitonov, Copet, Adi, Hsu, Elkahky,
  Tomasello, Algayres, Sagot, Mohamed, and Dupoux]{nguyen2023dgslm}
T.~A. Nguyen, E.~Kharitonov, J.~Copet, Y.~Adi, W.-N. Hsu, A.~Elkahky,
  P.~Tomasello, R.~Algayres, B.~Sagot, A.~Mohamed, and E.~Dupoux.
\newblock Generative spoken dialogue language modeling.
\newblock \emph{Transactions of the Association for Computational Linguistics},
  11:\penalty0 250--266, 2023.
\newblock URL \url{https://aclanthology.org/2023.tacl-1.15/}.

\bibitem[Ohashi et~al.(2026)Ohashi, Zeghidour, D{\'e}fossez, and
  Kharitonov]{ohashi2026interactivity}
A.~Ohashi, N.~Zeghidour, A.~D{\'e}fossez, and E.~Kharitonov.
\newblock Multi-faceted interactivity alignment in full-duplex speech models.
\newblock \emph{arXiv preprint arXiv:2606.11167}, 2026.
\newblock URL \url{https://arxiv.org/abs/2606.11167}.

\bibitem[{OpenAI}(2025)]{openai2025realtime}
{OpenAI}.
\newblock {GPT-Realtime} model.
\newblock OpenAI Platform documentation, 2025.
\newblock URL \url{https://developers.openai.com/api/docs/models/gpt-realtime}.
\newblock Accessed 2026-09-09.

\bibitem[Rezaei et~al.(2026)Rezaei, Mahmoud, Wang, Tyagi, Gosai, Dumitru,
  Sabharwal, Liu, and He]{rezaei2026rgsd}
M.~Rezaei, A.~Mahmoud, Z.~Wang, U.~Tyagi, A.~Gosai, R.-G. Dumitru,
  A.~Sabharwal, B.~Liu, and Y.~He.
\newblock Rubric-guided self-distillation: Post-training without rubric
  verifiers.
\newblock \emph{arXiv preprint arXiv:2606.12507}, 2026.
\newblock URL \url{https://arxiv.org/abs/2606.12507}.

\bibitem[Roy et~al.(2026)Roy, Raiman, Lee, Ene, Kirby, Kim, Kim, and
  Catanzaro]{roy2026personaplex}
R.~Roy, J.~Raiman, S.-g. Lee, T.-D. Ene, R.~Kirby, S.~Kim, J.~Kim, and
  B.~Catanzaro.
\newblock {PersonaPlex}: Voice and role control for full duplex conversational
  speech models.
\newblock \emph{arXiv preprint arXiv:2602.06053}, 2026.
\newblock URL \url{https://arxiv.org/abs/2602.06053}.

\bibitem[Sakshi et~al.(2025)Sakshi, Tyagi, Kumar, Seth, Selvakumar, Nieto,
  Duraiswami, Ghosh, and Manocha]{sakshi2025mmau}
S.~Sakshi, U.~Tyagi, S.~Kumar, A.~Seth, R.~Selvakumar, O.~Nieto, R.~Duraiswami,
  S.~Ghosh, and D.~Manocha.
\newblock {MMAU}: A massive multi-task audio understanding and reasoning
  benchmark.
\newblock In \emph{International Conference on Learning Representations}, 2025.
\newblock URL \url{https://arxiv.org/abs/2410.19168}.

\bibitem[Selvakumar et~al.(2025)Selvakumar, Seth, Anand, Tyagi, Kumar, Ghosh,
  and Manocha]{selvakumar2025multivox}
R.~Selvakumar, A.~Seth, N.~Anand, U.~Tyagi, S.~Kumar, S.~Ghosh, and D.~Manocha.
\newblock {MULTIVOX}: A benchmark for evaluating voice assistants for
  multimodal interactions.
\newblock In \emph{Proceedings of the 2025 Conference on Empirical Methods in
  Natural Language Processing}, pages 28481--28493, Suzhou, China, 2025.
  Association for Computational Linguistics.
\newblock \doi{10.18653/v1/2025.emnlp-main.1447}.
\newblock URL \url{https://aclanthology.org/2025.emnlp-main.1447/}.

\bibitem[Seth et~al.(2026)Seth, Kumar, Selvakumar, Anand, Tyagi, Seetharaman,
  Duraiswami, and Manocha]{seth2026aha}
A.~Seth, S.~Kumar, R.~Selvakumar, N.~Anand, U.~Tyagi, P.~Seetharaman,
  R.~Duraiswami, and D.~Manocha.
\newblock Audio hallucination attacks: Probing the reliability of large audio
  language models.
\newblock \emph{arXiv preprint arXiv:2603.29263}, 2026.
\newblock URL \url{https://arxiv.org/abs/2603.29263}.

\bibitem[Shao et~al.(2024)Shao, Wang, Zhu, Xu, Song, Bi, Zhang, Zhang, Li, Wu,
  and Guo]{shao2024deepseekmath}
Z.~Shao, P.~Wang, Q.~Zhu, R.~Xu, J.~Song, X.~Bi, H.~Zhang, M.~Zhang, Y.~Li,
  Y.~Wu, and D.~Guo.
\newblock {DeepSeekMath}: Pushing the limits of mathematical reasoning in open
  language models.
\newblock \emph{arXiv preprint arXiv:2402.03300}, 2024.
\newblock URL \url{https://arxiv.org/abs/2402.03300}.

\bibitem[Shi et~al.(2026)Shi, Wang, Guo, Wang, Zhang, Zhang, Guo, Hao, Xi,
  Yang, Xu, Zhou, and Lin]{shi2026qwen3asr}
X.~Shi, X.~Wang, Z.~Guo, Y.~Wang, P.~Zhang, X.~Zhang, Z.~Guo, H.~Hao, Y.~Xi,
  B.~Yang, J.~Xu, J.~Zhou, and J.~Lin.
\newblock {Qwen3-ASR} technical report.
\newblock \emph{arXiv preprint arXiv:2601.21337}, 2026.
\newblock URL \url{https://arxiv.org/abs/2601.21337}.

\bibitem[Tyagi et~al.(2026)Tyagi, Guo, Rezaei, George, Mahmoud, Lee, Liu, and
  He]{tyagi2026pow3r}
U.~Tyagi, X.~Guo, M.~Rezaei, D.~George, A.~Mahmoud, J.~Lee, B.~Liu, and Y.~He.
\newblock Not every rubric teaches equally: Policy-aware rubric rewards for
  {RLVR}.
\newblock \emph{arXiv preprint arXiv:2605.20164}, 2026.
\newblock URL \url{https://arxiv.org/abs/2605.20164}.

\bibitem[Veluri et~al.(2024)Veluri, Peloquin, Yu, Gong, and
  Gollakota]{veluri2024syncllm}
B.~Veluri, B.~N. Peloquin, B.~Yu, H.~Gong, and S.~Gollakota.
\newblock Beyond turn-based interfaces: Synchronous {LLMs} as full-duplex
  dialogue agents.
\newblock In \emph{Proceedings of the 2024 Conference on Empirical Methods in
  Natural Language Processing}, pages 21390--21402, 2024.
\newblock URL \url{https://aclanthology.org/2024.emnlp-main.1192/}.

\bibitem[Wang et~al.(2024)Wang, Lu, Tang, Yan, Xia, and
  Xiong]{wang2024fullduplex}
P.~Wang, S.~Lu, Y.~Tang, S.~Yan, W.~Xia, and Y.~Xiong.
\newblock A full-duplex speech dialogue scheme based on large language models.
\newblock In \emph{Advances in Neural Information Processing Systems}, 2024.
\newblock URL \url{https://arxiv.org/abs/2405.19487}.

\bibitem[Wu et~al.(2025)Wu, Mazar{\'e}, Zeghidour, and
  D{\'e}fossez]{wu2025aligning}
A.~Wu, L.~Mazar{\'e}, N.~Zeghidour, and A.~D{\'e}fossez.
\newblock Aligning spoken dialogue models from user interactions.
\newblock In \emph{International Conference on Machine Learning}, 2025.
\newblock URL \url{https://arxiv.org/abs/2506.21463}.

\bibitem[Zhang et~al.(2024)Zhang, Cheng, Deng, Chen, Wang, Zheng, Liu, Yu, Tan,
  Du, and Zhang]{zhang2024omniflatten}
Q.~Zhang, L.~Cheng, C.~Deng, Q.~Chen, W.~Wang, S.~Zheng, J.~Liu, H.~Yu, C.~Tan,
  Z.~Du, and S.~Zhang.
\newblock {OmniFlatten}: An end-to-end {GPT} model for seamless voice
  conversation.
\newblock \emph{arXiv preprint arXiv:2410.17799}, 2024.
\newblock URL \url{https://arxiv.org/abs/2410.17799}.

\bibitem[Z{\"u}fle et~al.(2026)Z{\"u}fle, Klejch, Sanders, Niehues, Birch, and
  Lam]{zufle2026factor}
M.~Z{\"u}fle, O.~Klejch, N.~Sanders, J.~Niehues, A.~Birch, and T.~K. Lam.
\newblock {F-Actor}: Controllable conversational behaviour in full-duplex
  models.
\newblock \emph{arXiv preprint arXiv:2601.11329}, 2026.
\newblock URL \url{https://arxiv.org/abs/2601.11329}.

\end{thebibliography}

\appendix
\small
\section{Training Hyperparameters}
\label{app:hparams}
Table~\ref{tab:hparams} gives the training settings. The supervised run uses a Moshi-style $7$B backbone, $80$ H100 GPUs, and $3{,}600$ steps; the reported SFT checkpoint is step $2{,}925$. Both RL stages leave model parameters trainable, with policy gradients flowing through the text head and shared temporal transformer. Decoded speech and transcripts supply rewards; audio-codebook actions receive no direct policy loss.

{
\footnotesize
\setlength{\tabcolsep}{4pt}
\begin{table}[t]
\centering
\caption{Training settings. RL stages inherit the preceding policy and freeze a copy as reference.}\label{tab:hparams}
\begin{tabularx}{\linewidth}{@{}l>{\raggedright\arraybackslash}X@{}}
\toprule
\textbf{Setting} & \textbf{Value} \\
\midrule

Base model & \textsc{Moshi}~\citep{defossez2024moshi} \\
SFT execution & $80$ H100 GPUs; batch $8$/GPU; global batch $640$ \\
SFT budget & $3{,}600$ steps; $2.304$M fixed sample draws \\
Reported model checkpoint & step $2{,}925$ ($1.872$M draws) \\
SFT optimizer & AdamW; lr $2.828{\times}10^{-6}$; wd $0.1$ \\
Depth-former lr & $5.657{\times}10^{-6}$ \\
SFT warmup / gradient clip & $500$ steps / $3.0$ \\
Audio/text sampling mass & $0.88/0.12$ \\
First-codebook / text-pad weight & $100/0.5$ \\
Turn / backchannel onset weight & $1.5/3.0$ \\
Context and response ceiling & $300$\,s \\
RL learning rate & $5{\times}10^{-7}$ \\
RL algorithm & component-normalized GDPO (weighted sum; final batch normalization) \\
RL KL coefficient / target & $0.05/0.01$ (adaptive, minimum coefficient $0.02$; bounded k3 estimator; both stages) \\
RL clip / gradient clip & $0.2/1.0$ \\
RL response-continuity term & weight $0.5$, target $4.0$\,s (both stages) \\
RL continuation-duration bonus (stage 2) & weight $2.0$, target $4$\,s, on noise-robustness and user-backchannel events \\
RL reward weights & interactivity $1.0$; transcript rubric judge $0.75$ (Gemini 3.6 Flash) \\
RL event context / max response & $30$\,s / $30$\,s \\
RL policy stream & text stream incl.\ padding actions \\
Parameter dtype & bfloat16 (fp32 master persisted for RL); weight-delta gate ($\geq5\%$ of elements changed, median $\geq1$ ULP) in both stages \\
Gradient checkpointing & on \\
\bottomrule
\end{tabularx}
\end{table}
}

\subsection{Reward Weights}
\label{app:reward-weights}
Table~\ref{tab:reward-weights} separates rewards with within-group variation from hard validity checks and held-out evaluation metrics. In the stage-1 run the transcript rubric judge is active on the turn and interruption strata only, with within-group standard deviations of $0.14$--$0.43$ and no parse failures in the recorded rollouts.

{
\small
\setlength{\tabcolsep}{3pt}
\begin{table}[t]
\centering
\caption{Reward components and evaluation roles in the reported RL run (interactivity-v2 with the response-continuity term; stage 2 adds the continuation-duration bonus). Component rewards are normalized within rollout groups; hard validity checks and held-out metrics never substitute for missing reward variation.}\label{tab:reward-weights}
\begin{tabularx}{\linewidth}{@{}l>{\raggedright\arraybackslash}X@{}}
\toprule
\textbf{Component} & \textbf{Weight} \\
\midrule
Pause timing & component reward ($1.0$) \\
Turn timing + transcript rubric judge & component rewards ($1.0$ timing; $0.75$ rubric, pre-boundary transcript shown) \\
Backchannel timing & component reward ($1.0$) \\
Response continuity (first response) & component term ($0.5$, target $4.0$\,s) on turn, interruption, paired pause, and paired backchannel groups \\
Continuation duration (stage 2) & component term ($2.0$, target $4$\,s) on noise-robustness and user-backchannel groups \\
Noise robustness / speech mirror & component rewards ($1.0$) \\
Interruption yield + recovery & component rewards ($1.0$ each) \\
Waveform integrity & hard rollout validity \\
Malformed or incomplete judge result & invalid rollout \\
Safety and broad capability & held-out evaluation \\
AudioMC / \textsc{SteerBench} / VoiceBench & frozen-checkpoint evaluation \\
\bottomrule
\end{tabularx}
\end{table}
}

\subsection{Compute Resources}
\label{app:compute}
The first RL stage ran on four nodes with eight H100 80\,GB GPUs each; the second used one eight-GPU H100 80\,GB node. From run initialization to saving the checkpoints used in the paper, the recorded intervals were $5.08$ and $4.01$ hours, respectively, corresponding to approximately $162.7$ and $32.1$ allocated GPU-hours. These estimates include rollout generation, reward computation, optimization, and checkpoint writes within each interval. They exclude queueing, prior setup, benchmark evaluation, and hosted-model compute. Preliminary runs and training beyond these checkpoints used additional compute, so this subtotal does not represent the full project.

\newpage
\section{Evaluation and Checkpoint Selection}
\label{app:protocol}
\label{sec:protocol-prereg}
\subsection{Judges and Checkpoint Selection}
SteerBench and FDB-v2 use Gemini 3.6 Flash. AudioMC, VoiceBench, and FDB-v1 interruption ratings use gpt-5.4-mini at medium reasoning effort; VoiceBench uses three judge samples per item. FDB-v1/v2 transcription uses parakeet-tdt-0.6b-v2. Judge identity and scoring conventions are checked before aggregation. The judge-sensitivity study uses the same generations under both judges; it does not change the reported checkpoint.

The development set is separate from benchmark test sets and targets the same capabilities. We select and freeze the checkpoint using development performance alone, then report benchmark results.

\subsection{Overlap and Pause Measurement}
An overlap check found strong transcript $n$-gram overlap between $100$ of $216$ CANDOR pause transcripts and $96$ supervised conversations. CANDOR turn and pause tasks are therefore diagnostic for the entire training lineage. Official pause clips also end only $0.02$--$0.11$ seconds after the user's last word, so they cannot measure successful waiting followed by a response. The synthetic pause probe supplies the continuation needed for that measurement. The source-clean FDB-v1.5 comparison uses $498$ examples.

\begin{samepage}
\section{RL Controls and Robustness}\label{app:ablations}
\subsection{Complete Retention Results}
Table~\ref{tab:retention-details} reports three-run means, population standard deviations, and differences for the compact main-text comparison. It also includes semantic quality, safety, and AudioMC ARS.
\end{samepage}
\begin{table}[t]
\centering
\caption{\textbf{Capability comparison.} Means $\pm$ population standard deviations over three runs per model and benchmark. The same runs supply every subscore. Higher is better; bold marks the larger displayed mean, including ties. Differences are computed before rounding. Rates and their differences use percentages and percentage points; other scores retain their native scales.}
\label{tab:retention-details}
\small
\setlength{\tabcolsep}{6pt}
\begin{tabular}{@{}lrrr@{}}
\toprule
Metric & SFT & + RL & $\Delta$ \\
\midrule
AudioMC APR (\%) & 13.64$\pm$0.28 & \textbf{14.38$\pm$0} & $+0.74$ \\
AudioMC ARS (\%) & 37.17$\pm$1.33 & \textbf{37.97$\pm$0.44} & $+0.80$ \\
Interruption response (\%) & 96$\pm$0.4 & \textbf{97.7$\pm$0.2} & $+1.67$ \\
Interruption semantics & \textbf{3.94$\pm$0.03} & 3.88$\pm$0.03 & $-0.069$ \\
FDB-v2 mean & \textbf{4.17$\pm$0.02} & \textbf{4.17$\pm$0.03} & $-0.003$ \\
FDB-v2 safety & 4.65$\pm$0.05 & \textbf{4.81$\pm$0.05} & $+0.153$ \\
FDB-v2 turn taking & 4.18$\pm$0.13 & \textbf{4.19$\pm$0.07} & $+0.018$ \\
FDB-v2 instruction following & 3.67$\pm$0.13 & \textbf{3.81$\pm$0.26} & $+0.134$ \\
\textsc{SteerBench} rubric pass (\%) & 63.75$\pm$1.59 & \textbf{65.22$\pm$0.47} & $+1.47$ \\
VoiceBench overall & 40.87$\pm$0.27 & \textbf{41.38$\pm$0.02} & $+0.51$ \\
\bottomrule
\end{tabular}

\end{table}

\subsection{Training Controls}
\begin{table}[t]
\centering
\caption{\textbf{RL controls.} Differences between each model's three-run mean and the corresponding SFT mean. Higher is better; bold marks the largest point estimate per column. Data and training-budget differences prevent interpreting these as single-variable ablations.}
\label{tab:rl-arms}
\footnotesize
\setlength{\tabcolsep}{3pt}
\begin{tabular}{@{}lccc@{}}
\toprule
Variant & FDB-v2 mean & FDB-v2 turn taking & VoiceBench \\
\midrule
Original timing recipe & -0.115 & -0.244 & +0.049 \\[3pt]
Stage 1 & +0.107 & -0.140 & +0.444 \\[3pt]
\rowcolor{scaleBlue!8}
\textsc{SteerDuplex}-RL & -0.003 & +0.018 & \textbf{+0.505} \\[3pt]
Training-seed replicate & \textbf{+0.155} & \textbf{+0.048} & +0.227 \\[3pt]
Single-stage & -0.177 & -0.050 & -0.363 \\[3pt]
\bottomrule
\end{tabular}
\end{table}

The original-recipe control starts from SFT with the original interaction recipe and no continuity additions. Stage 1 starts from SFT with continuity weight $0.5$ and a $4$-second target; its trained checkpoint initializes stage 2. Stage 2 adds a continuation bonus (weight $2.0$, target $4$ seconds) and dedicated user-backchannel sampling. The single-stage variant applies the final reward directly from SFT with a smaller training budget. The training-seed replicate changes the random seed of the reported configuration. These controls differ in training budget and, for the original recipe, additional configuration choices, so they do not isolate every reward component.

\begin{table}[t]
\centering
\caption{\textbf{Judge sensitivity.} FDB-v2 differences on identical generations from three runs per model; higher is better. Columns use Gemini 3.6 Flash (canonical) and gpt-5.4-mini (low reasoning). Both judges score the same runs from the common rescored pool. The broader comparison is in Table~\ref{tab:retention-details}.}
\label{tab:crossjudge-fdb2}
\footnotesize
\setlength{\tabcolsep}{5pt}
\begin{tabular}{@{}llrr@{}}
\toprule
Comparison & Metric & Gemini & GPT \\
\midrule
Stage 1 minus SFT & headline mean & $+0.057$ & $-0.058$ \\
Stage 1 minus SFT & per-event turn taking & $-0.278$ & $-0.106$ \\
Stage 1 minus SFT & per-event instruction following & $-0.065$ & $-0.009$ \\
RL minus SFT & headline mean & $-0.083$ & $-0.098$ \\
RL minus SFT & per-event turn taking & $-0.074$ & $-0.059$ \\
RL minus SFT & per-event instruction following & $-0.124$ & $+0.001$ \\
\bottomrule
\end{tabular}
\end{table}

\subsection{Reward and Continuation Snapshots}
\label{app:reward-dynamics}
Table~\ref{tab:reward-dynamics} lists the snapshots in Figure~\ref{fig:reward-integrity}. Training and development observations were recorded separately within one run; no intermediate measurements or confidence intervals are available.
\begin{table}[t]
\centering
\caption{\textbf{Second-stage optimization snapshots.} Each row compares earlier and later observations on its own scale. The SFT continuation reference is $2.60$ seconds.}
\label{tab:reward-dynamics}
\small
\begin{tabular}{@{}lrr@{}}
\toprule
Measurement & Earlier & Later \\
\midrule
Training noise-robustness reward & 1.96 & 0.77 \\
Development interruption reward & 0.450 & 0.793 \\
Continuation after user backchannel (s) & 3.20 & 2.00 \\
\bottomrule
\end{tabular}
\end{table}

\FloatBarrier
\begin{samepage}
\section{Prompts and Judges}\label{app:prompts}\subsection{System Prompts}
\label{app:system-prompt}
This section lists the two prompt templates used by \textsc{SteerDuplex}, the constant training system prompt and the shared inference system prompt, and describes the multi-turn context wrapper that surfaces dialogue history.
\end{samepage}

The training-time system prompt below is prepended to the agent-text channel of every RL rollout; per-scenario steering instructions are inserted on the user side via the dialogue history.
\begin{quote}\itshape\small
You are a helpful full-duplex voice assistant. Your voice and identity are fixed. Stay kind, safe, and constructive. Follow the user's spoken instructions about tone, persona flavor, speaking style, speed, and length when those instructions are allowed.
\end{quote}

\needspace{5\baselineskip}
For inference and benchmark evaluation, the shared default assistant prompt is:
\begin{quote}\itshape\small
You are a helpful voice assistant. Your voice and identity are fixed. Listen carefully, follow the user's instructions, and respond naturally.
\end{quote}

For multi-turn RL scenarios with stitched user audio, the system prompt includes a plain ``Conversation so far'' transcript. It alternates user and assistant lines, marks assistant-generated text, and ends with the latest user turn. The model is instructed to respond to that turn using the preceding context.

\Needspace{28\baselineskip}
\subsection{Reference-Audio Rubric Judge}
\label{app:audio-rubric-prompt}
The reference-audio rubric judge (§\ref{sec:ref-rewards}) receives two clips (Audio~1: target style; Audio~2: model output) and their transcripts, then returns per-criterion JSON scores. The template below is simplified for presentation; the executable prompt accompanies the evaluation code.
\begin{promptlisting}
You are judging audio steerability for a full-duplex voice assistant.
Audio 1 is the reference target.
Audio 2 is the model-generated answer.
Compare delivery / style / prosody / speaking rate / articulation / affect against the reference. Do NOT require identical speaker timbre or voice identity unless the rubric explicitly asks for it. Use the transcript only to verify content; judge audio qualities from the audio.
Scenario category: <category>
User request/context: <user_text>
Reference transcript: <ref_text>
Generated transcript: <gen_text>
Audio rubrics:
<numbered list of rubric items>
Return ONLY valid JSON with numeric scores in [0,1]:
{"reason": "<one sentence>", "score": 0.0,
 "style_match": 0.0, "prosody_match": 0.0, "speech_rate_length": 0.0,
 "articulation": 0.0, "naturalness": 0.0, "task_success": 0.0,
 "criteria": [{"title": "...", "category": "...",
               "rating": "No Issues|Minor Issues|Major Issues",
               "score": 0.0, "reason": "..."}]}
\end{promptlisting}

\begin{table}[t]
\centering
\small
\setlength{\tabcolsep}{4pt}
\caption{\textbf{Single-family reward probes vs.\ the joint composite.} Endpoint
diagnostics on $96$ held-out sampled rollouts per condition. Scalar rewards use different objectives; the common behavioral diagnostics show that every isolated probe leaves the empty-rollout rate
$\geq 26\%$ and the duplex-FDB score at or near zero ($\leq 0.055$).}
\label{tab:probe-summary}
\begin{tabular}{lcccc}
\toprule
\textbf{Probe} & \textbf{Scalar$\uparrow$} & \textbf{Empty/96$\downarrow$} & \textbf{Duplex$\uparrow$} & \textbf{MOS$\uparrow$} \\
\midrule
Text-rubric only      & 0.329 & 47 & 0.055 & 0.512 \\
Audio-rubric only     & 0.336 & 41 & 0.000 & 0.571 \\
Audio-quality only    & 0.522 & 41 & 0.000 & 0.561 \\
Promptness only       & 0.670 & 25 & 0.000 & 0.626 \\
\midrule
Joint composite probe & 0.554 & \textbf{4} & \textbf{0.371} & \textbf{0.631} \\
\bottomrule
\end{tabular}
\end{table}

\Needspace{23\baselineskip}
\subsection{Benchmark Judge}
\label{app:benchmark-judge-prompt}
For benchmark-aligned scenarios (daily, correction, safety, entity-tracking, factual, AudioMC), the benchmark judge is conditioned with the following system prompt and a per-task rubric (one of: daily, correction, entity-tracking, safety, alpaca, factual, ifeval, default):
\begin{promptlisting}
You judge spoken-assistant transcripts. Given the user's request, any scenario-specific rubric, and the assistant's transcript, FIRST write a ONE-SENTENCE explanation of how the assistant performed (referencing concrete details from the transcript), THEN give a 1..5 integer score (5 = best). If a list of criteria is provided, evaluate each one independently with a one-sentence reason and a boolean flag.
Return ONLY valid JSON with this exact key ORDER (reasoning first, scores last) and no other text:
{"reason": "<ONE sentence, <= 220 chars>",
 "criteria_reasons": ["...", ...], "criteria_met": [<bool>, ...],
 "task_score": <int 1..5>}
Be calibrated. A typical OK-but-imperfect response is a 3. A 5 requires the response to actually accomplish the task as a competent human would. Always reason FIRST, score AFTER - never invert the order.
\end{promptlisting}

\Needspace{12\baselineskip}
\subsection{Text Rubric Judge}
\label{app:text-rubric-prompt}
The text rubric judge (Section~\ref{sec:rewards-design}) scores one criterion at a time as ``No Issues,'' ``Minor Issues,'' or ``Major Issues,'' explaining its decision before rating. It receives criterion metadata (title, category, type, weight, and description), the latest user request, and the assistant transcript. Explicit criteria require direct answers; implicit criteria may be inferred from context. Objective criteria concern factual correctness, while subjective criteria concern quality. The judge ignores criterion weights and tolerates minor formatting or phrasing differences. For speech, content takes priority over delivery polish. Output is JSON with reasoning before the rating.

\subsection{Agreement with Human Annotations}
\label{app:human-agreement}
We assess reference-audio judging by comparing $200$ LLM-generated labels with $200$ corresponding human annotations. Agreement, measured as the fraction of matching labels, is approximately $76\%$ with reference audio and $62\%$ without it. The approximately $14$-percentage-point gain supports reference conditioning in this comparison, while agreement remains imperfect.

\section{Reward-Integrity Diagnostic}
\label{app:reward-hacking-details}
\label{app:matrix-kl}
The diagnostic compares four isolated reward families with a joint composite on $96$ held-out sampled rollouts per condition (Table~\ref{tab:probe-summary}). The composite includes semantic, audio, quality, interaction, instruction-following, safety, and promptness components. These small probes are separate from the reported RL checkpoint. Their endpoint scores establish failure modes within this study, not a general necessity result for a particular reward combination.

\Needspace{7\baselineskip}
\section{Baseline Systems}
\label{app:baselines}
\subsection{Configurations}
\paragraph{Open models.}
We use \textsc{Moshi}'s public \textit{moshiko-pytorch-bf16} checkpoint~\citep{defossez2024moshi} and \textsc{PersonaPlex}'s \textit{personaplex-7b-v1} checkpoint through its official inference server~\citep{roy2026personaplex}. Both are $7$B full-duplex speech models.

\paragraph{Proprietary context.}
The AudioMC comparison reproduces Gemini 2.5 Flash with text output and GPT Realtime with audio output from \citet{gosai2025audiomc}. GPT Realtime (\texttt{gpt-realtime}) is distinct from the older GPT-4o Realtime model~\citep{openai2025realtime}. VoiceBench reports separate hosted-API comparisons; Gemini Live refers to Google's streaming API~\citep{google2025geminilive}. Imported scores retain their original evaluation protocols.

\FloatBarrier
\subsection{Benchmark Breakdowns}
Tables~\ref{tab:steerability} and~\ref{tab:fdb_v15} provide the exact values behind Figures~\ref{fig:steerbench} and~\ref{fig:overlap}.
\begin{table}[t]
\centering
\caption{\textbf{Exact \textsc{SteerBench} results underlying Figure~\ref{fig:steerbench}.} Matched items, rubric implementation, and Gemini 3.6 Flash judge. SFT averages three runs; $\pm$ denotes population standard deviation. Audio APR requires every audio rubric to pass; sample APR requires every text and audio rubric to pass. Rubric pass rate averages individual decisions. All values are percentages; higher is better. Bold marks the best displayed score in each column.}
\small
\noindent\begin{minipage}[c]{0.37\textwidth}
\centering
\textbf{Matched system comparison}\\[3pt]
\setlength{\tabcolsep}{6pt}
\begin{tabular}{@{}lr@{}}
\toprule
\textbf{System} & \textbf{Audio APR (\%)} \\
\midrule
\textsc{Moshi}~\citep{defossez2024moshi} & 20.55 \\
\textsc{PersonaPlex}~\citep{roy2026personaplex} & 16.44 \\
\rowcolor{scaleBlue!8}
\textsc{SteerDuplex} (ours) & \textbf{65.10$\pm$1.13} \\
\bottomrule
\end{tabular}
\end{minipage}
\hfill
\begin{minipage}[c]{0.58\textwidth}
\centering
\textbf{\textsc{SteerDuplex} steering profile}\\[3pt]
\setlength{\tabcolsep}{6pt}
\begin{tabular}{lrr}
\toprule
\textbf{Axis} & \textbf{Sample APR (\%)} & \textbf{Rubric pass (\%)} \\
\midrule
\rowcolor{scaleBlue!8}
Tone & 49.67 & 71.51 \\
\rowcolor{scaleBlue!8}
Persona & \textbf{51.11} & \textbf{77.03} \\
\rowcolor{scaleBlue!8}
Style / accent & 48.40 & 69.83 \\
\rowcolor{scaleBlue!8}
Speed / length & 32 & 63.10 \\
\bottomrule
\end{tabular}
\end{minipage}
\label{tab:steerability}
\end{table}

\begin{table}[t]
\centering
\caption{\textbf{Source-clean interruption and overlap handling (FDB-v1.5).} Success rates (\%) on $498$ paired examples. Bold marks the highest success rate in each column. All checkpoints use the frozen behavior scorer and Gemini 3.6 Flash judge. The stage-1 row is behavior-only because its user-backchannel acoustic check was unavailable. Other checkpoints pass silence, clipping, and duration checks.}
\label{tab:fdb_v15}
\small
\setlength{\tabcolsep}{6pt}
\begin{tabular}{@{}lrrrr@{}}
\toprule
 & Interruption & Background & Talking to & User \\
Checkpoint & response & recovery & another & backchannel \\
\midrule
Examples & 200 & 100 & 100 & 98 \\
\rowcolor{scaleBlue!8}
SFT & 72.5 & \textbf{60} & 42 & 71.4 \\
\rowcolor{scaleBlue!8}
RL stage 1 & 80.5 & 53 & 45 & 65.3 \\
\rowcolor{scaleBlue!8}
+ RL & \textbf{82.5} & 59 & \textbf{48} & \textbf{80.6} \\
\bottomrule
\end{tabular}
\end{table}

Tables~\ref{tab:fdb_v2}, \ref{tab:audiomc}, and~\ref{tab:voicebench} supplement Table~\ref{tab:capabilities} with expanded RL comparisons and hosted-API context.
\begin{table}[t]
\centering
\captionsetup{hypcap=false}
\captionof{table}{\textbf{Expanded FDB-v2 task comparison.} Slow-examiner scores on $200$ sessions per pass, judged by Gemini 3.6 Flash ($1$--$5$; higher is better). Each model averages three runs; the same runs supply all categories. Bold marks the higher displayed score (both for ties).}\label{tab:fdb_v2}
\footnotesize
\setlength{\tabcolsep}{6pt}
\begin{tabular}{lccccc}
\toprule
\textbf{System} & \textbf{Corr.} & \textbf{Daily} & \textbf{Entity} & \textbf{Safety} & \textbf{Mean} \\
\midrule
\multicolumn{6}{l}{\textit{Three runs per model}} \\
\rowcolor{scaleBlue!8}
\textsc{SteerDuplex} (matched ref.) & 4.21 & \textbf{3.74} & \textbf{4.07} & 4.65 & \textbf{4.17} \\
\rowcolor{scaleBlue!8}
\quad + RL & \textbf{4.41} & 3.41 & 4.04 & \textbf{4.81} & \textbf{4.17} \\
\bottomrule
\end{tabular}

\end{table}
\begin{table}[t]
\centering
\captionsetup{hypcap=false}
\captionof{table}{\textbf{Hosted-API AudioMC context.} Scores from \citet{gosai2025audiomc}, in percent; higher is better. Gemini 2.5 Flash uses text output; GPT Realtime uses audio output. Axes follow Table~\ref{tab:capabilities}. Bold marks the best displayed score. Different protocols limit comparisons with open models.}\label{tab:audiomc}
\footnotesize
\setlength{\tabcolsep}{5pt}
\begin{tabular}{lcccccc}
    \toprule
    \multirow{2}{*}{\textbf{System}}
        & \multicolumn{2}{c}{\textbf{Overall}}
        & \multicolumn{4}{c}{\textbf{Per-Axis APR (\%)}} \\
    \cmidrule(lr){2-3} \cmidrule(lr){4-7}
        & APR (\%) & ARS (\%) & IM & IR & SC & VE \\
    \midrule
    \multicolumn{7}{l}{\textit{Proprietary}} \\
    \textcolor{gray}{Gemini 2.5 Flash}
        & \textcolor{gray}{\textbf{26.11}} & \textcolor{gray}{\textbf{65.42}}
        & \textcolor{gray}{\textbf{19.70}} & \textcolor{gray}{\textbf{29.17}}
        & \textcolor{gray}{\textbf{31.33}} & \textcolor{gray}{\textbf{26.50}} \\
    \textcolor{gray}{GPT Realtime}
        & \textcolor{gray}{20.35} & \textcolor{gray}{63.03}
        & \textcolor{gray}{\textbf{19.70}} & \textcolor{gray}{20}
        & \textcolor{gray}{26.51} & \textcolor{gray}{17.09} \\
    \bottomrule
\end{tabular}

\end{table}
\begin{table}[t]
\centering
\caption{\textbf{Additional VoiceBench scores.} Gray rows provide hosted-API context. The matched comparison averages three runs per model, with three judge samples per item. The subscore breakdown includes only the judged SD-QA score. Bold marks the best displayed subscore within each comparison block, including ties.}
\scriptsize
\setlength{\tabcolsep}{3.2pt}
\begin{tabular}{lcccccccc}
\toprule
\textbf{Model}
    & \textbf{AlpacaEval}
    & \textbf{CommonEval}
    & \textbf{WildVoice}
    & \textbf{SD-QA}
    & \textbf{IFEval}
    & \textbf{BBH}
    & \textbf{AdvBench}
    & \textbf{Overall} \\
\midrule
\multicolumn{9}{l}{\textit{Proprietary}} \\
\textcolor{gray}{Gemini Live}
    & \textcolor{gray}{3.69} & \textcolor{gray}{3.52} & \textcolor{gray}{3.26}
    & \textcolor{gray}{53.5 / 40.6}
    & \textcolor{gray}{17 / 25.9}
    & \textcolor{gray}{62.8} & \textcolor{gray}{94.1} & \textcolor{gray}{63.01} \\
\textcolor{gray}{GPT-4o Realtime}
    & \textcolor{gray}{\textbf{4.65}} & \textcolor{gray}{\textbf{4.33}} & \textcolor{gray}{\textbf{4.38}}
    & \textcolor{gray}{\textbf{88.9 / 75.9}}
    & \textcolor{gray}{\textbf{24.3 / 37.3}}
    & \textcolor{gray}{\textbf{62.9}} & \textcolor{gray}{\textbf{98.8}} & \textcolor{gray}{\textbf{78.42}} \\
\midrule
\multicolumn{9}{l}{\textit{Three runs per model}} \\
\rowcolor{scaleBlue!8}
\textsc{SteerDuplex} (matched ref.) & 2.28 & 2.21 & 1.74 & 26.88 / -- & -- & -- & \textbf{99.3} & 40.87 \\
\rowcolor{scaleBlue!8}
\quad + RL & \textbf{2.44} & \textbf{2.28} & \textbf{1.77} & \textbf{28.27} / -- & -- & -- & 99.2 & \textbf{41.38} \\
\bottomrule
\end{tabular}
\label{tab:voicebench}
\end{table}

\section{Data and Responsible Development}
\label{app:data-stats}
The supervised mixture contains $504{,}416$ audio and $65{,}675$ text records, with $8{,}510.9$ hours of audio records and a fixed $2.304$M sample-draw budget. Audio and text have sampling masses $0.88$ and $0.12$. Targeted subsets include $20{,}000$ multi-turn instruction-following records from $5{,}000$ sessions, $25{,}000$ steering records from $5{,}000$ sessions, $1{,}200$ duplex records, and $1{,}300$ six-turn mathematics conversations. Dataset splits are fixed at the record-list level; duplicate paths are removed and long examples are split at turn boundaries. The duration total counts training records, not unique source recordings.

\paragraph{Interaction RL data.}
The proprietary interaction pool covers turns, interruptions, pauses, backchannels, noise, and speech-mirror scenarios. CANDOR filtering removes identified overlap with evaluation conversations; official test prompts and labels are excluded from optimization. The second stage emphasizes continued speech after listener feedback through dedicated user-backchannel sampling.

\subsection{Contributor Consent and Data Handling}
\label{app:data-protocol}
Contributors received training on task design, rubric writing, audio quality, allowed content, and speaker consent. They submitted prompts, recordings, transcripts, reference answers, and rubrics for independent review. Screening rejected audio corruption, transcript mismatches, personal information, inappropriate content outside designated safety tasks, and missing speaker permission.

Recordings were restricted to contributors' own voices or explicitly consenting participants. Instructions prohibited bystander speech, identifying personal information, and imitation of private individuals or public figures; fictional entities were used where needed. Contributors were paid $\$25$--$\$40$ per hour, and were informed of training, evaluation, and consent-dependent research release uses. The collection protocol obtained the applicable permissions and ethics approval. Raw audio release follows contributor consent and source terms.

\subsection{Demographic and Linguistic Coverage}
\label{app:demographics}
We evaluate English spoken dialogue only. We collect speaker-gender and native-accent metadata when contributors voluntarily provide it, and we use those fields for aggregate coverage checks rather than for individual profiling. AudioMC metadata covers $2{,}998$ user speakers with a $51.3\%$ female, $47.1\%$ male, and $1.6\%$ other gender split, and includes General American, African (non-South African), British, Canadian, Southern American, Northeastern U.S., Midwestern U.S., European, Indian, East Asian, Australian, Latin American, Middle Eastern, New Zealand, and Irish accent labels. We do not claim coverage of non-English languages, code-switching, children, clinical speech, or all regional dialects.

\subsection{Artifact Licenses and Terms}
\label{app:artifact-licenses}
Table~\ref{tab:artifact-licenses} summarizes the terms governing training and evaluation artifacts. Restricted corpora and benchmark assets are not redistributed with the paper.

\begin{table}[t]
\centering
\scriptsize
\renewcommand{\arraystretch}{0.92}
\setlength{\tabcolsep}{3pt}
\begin{tabularx}{\textwidth}{>{\raggedright\arraybackslash}p{2.35cm}
  >{\raggedright\arraybackslash}p{2.05cm}
  >{\raggedright\arraybackslash}p{3.15cm}
  >{\raggedright\arraybackslash}X}
\toprule
\textbf{Artifact} & \textbf{Role} & \textbf{License / terms} & \textbf{Use in this work} \\
\midrule
\textsc{SteerBench} & New benchmark & Research license; release upon acceptance & Steerability evaluation; evaluation prompts and labels excluded from optimization; audio release follows contributor consent terms. \\
Synthetic SFT data & Training & Author-generated internal artifact & SFT; release limited to artifacts whose source-data and generated-audio terms permit redistribution. \\
In-house two-person conversations & Training & Contributor agreement and consent terms & Training after quality, PII, and speaker-permission checks; raw recordings not redistributed without explicit permission. \\
Proprietary duplex development data & RL training & Project-specific data-use permission & Gradient-side development examples only; benchmark test material excluded; raw audio not redistributed. \\
Fisher English~\citep{cieri2004fisher} & Training & LDC User Agreement & Licensed conversational speech; not redistributed. \\
Audio MultiChallenge~\citep{gosai2025audiomc} & Evaluation / analysis & Open-source release; contributor consent terms & Multi-turn spoken evaluation and coverage analysis. \\
Full-Duplex-Bench v1/v2~\citep{lin2025fdbv1,lin2026fdbv2} & Evaluation & Public research release; no third-party redistribution & Duplex interaction metrics through the official harnesses. \\
VoiceBench~\citep{chen2024voicebench} & Evaluation & Apache-2.0 dataset/code release & Spoken instruction following, reasoning, safety, and QA evaluation. \\
\textsc{Moshi}~\citep{defossez2024moshi} & Base model / baseline & CC BY 4.0 model release & Base architecture and open-weight baseline. \\
\textsc{PersonaPlex}~\citep{roy2026personaplex} & Baseline & NVIDIA Open Model License; CC BY 4.0 additional information & Open-weight full-duplex baseline. \\
OpenAI / Google hosted models & Baselines & Provider API terms & Hosted APIs; no model weights redistributed. \\
gpt-5.4-mini / Gemini 3.6 Flash~\citep{google2026gemini36flash} & Judges & Provider API terms & Hosted APIs; scores only; no model outputs redistributed. \\
\bottomrule
\end{tabularx}
\caption{Licenses and use terms for major artifacts. When a third-party artifact has more restrictive redistribution terms than research use terms, we report results but do not redistribute the underlying asset.}
\label{tab:artifact-licenses}
\end{table}

\subsection{Misuse Risks}
\label{app:ethical-risks}
Steerable dialogue models can enable impersonation, manipulation, or unsafe persuasion. Training excludes non-consensual voice cloning, and the system prompt specifies a fixed assistant identity. Safety rubrics and refusal data reduce but do not eliminate misuse risks; deployment still requires identity, consent, abuse-monitoring, and content-safety controls. Our English-only evaluation does not establish safety across languages or accents.

\subsection{Writing Assistance}
\label{app:ai-disclosure}
The authors used AI tools for language editing, LaTeX cleanup, and writing assistance. They reviewed and edited all assisted text and take responsibility for the paper's research, claims, correctness, originality, and integrity.

\section{SteerBench Documentation}
\label{app:benchmark-documentation}
\paragraph{Contents and intended use.}
The evaluation set contains $100$ tone, $136$ persona, $104$ style/accent, and $50$ speed/length prompts. Items record identifiers, steering categories, subcategories, topics, user utterances, and audio/text rubrics. Rubrics pair an axis with a binary criterion; item identifiers link user audio, reference audio, and transcripts. The benchmark measures single-turn English steering under explicit requests; it does not test long-term personalization or coverage across languages.

\paragraph{Synthetic audio construction.}
User utterances are rendered as neutral, conversational speech with Gemini 3.1 Flash TTS at $24$\,kHz, mono, PCM16. A hash of the item identifier chooses a voice from a fixed pool, so the voice assignment is stable. For synthetic references, a text model converts the user request, steering category, and rubrics into a transcript and delivery instructions. The saved generation records identify Gemini 3 Pro Preview for the original scripts and Gemini 3.1 Pro Preview for repairs; synthesis uses Gemini 3.1 Flash TTS with Gemini 2.5 Flash TTS as a fallback. Reference voices are chosen deterministically from Kore and Schedar. Human review validates the requested delivery. Fixed reference clips are reused across models; regeneration can change the target.

\paragraph{Scoring and error handling.}
Text and audio criteria are judged separately, with audio decisions conditioned on the fixed reference. The scoring implementation excludes items with judge-call errors from pass-rate denominators and records their count; a run with more than $5\%$ such errors is invalid. Audio APR requires all audio criteria for an item to pass, sample APR requires all applicable criteria to pass, and rubric pass rate counts individual decisions. These definitions apply within each reported category as well as to the pooled set.

\paragraph{Access and reuse.}
Code, \textsc{SteerBench}, and checkpoints will be released upon acceptance under a research license. Audio release follows consent and source terms (Appendix~\ref{app:artifact-licenses}).

\subsection{Synthetic User-Audio Template}
The literal request template is reproduced below; \texttt{utterance} is the item's user request.
\begin{lstlisting}[style=steerprompt,basicstyle=\ttfamily\fontsize{8}{9}\selectfont,literate={—}{{\textemdash}}1]
Read the following user utterance naturally, like a real person talking to a voice assistant.

# Audio Profile
A real human user speaking casually to a voice assistant.

# Director's note
Style: Natural. Pace: Conversational. Accent: American (Gen).

## Scene:
A user speaking into a phone or laptop microphone.

## Transcript:
{utterance}
\end{lstlisting}

\subsection{Synthetic Reference-Script Template}
The generator uses the following system message. The user message supplies the item identifier, category, subcategory, user utterance, and rubric list, as shown afterward. The request specifies JSON output; the generator does not set an explicit sampling temperature.
\begin{lstlisting}[style=steerprompt,basicstyle=\ttfamily\fontsize{8}{9}\selectfont,literate={—}{{\textemdash}}1]
You generate REFERENCE audio scripts for a voice-assistant steerability benchmark.

For each bench sample you receive: a user utterance, the steering sub_axis the
model is supposed to exhibit, and the rubric items the model will be scored on.

You output a short response that an *ideal* steerable assistant would produce
for this turn — content that engages the user, in the requested style — plus a
director's note describing exactly how the line should be spoken.

The script will be sent to a TTS model. Your transcript and director's note
together MUST produce audio that clearly exhibits the target sub_axis. The
content is secondary — the audio's job is to anchor STYLE/TONE/PERSONA/ACCENT/
PACE for a judge model.

Style rules by split:

- A3_tone_controlled: the sub_axis is a TONE (angry, playful, sad, etc.).
  Transcript should be plain, natural prose engaging the user's topic.
  DO NOT add bracket prefixes. Put the entire tone signal in `tts_instruct` as
  a vivid director's note (e.g. "Sharp, irritated, clipped pacing; voice tight
  with frustration; short bitten-off phrases.").

- A4_persona_controlled: the sub_axis is a PERSONA (pirate_captain, doctor,
  noir_detective, etc.). Transcript SHOULD start with the inline tag
  `[like a <persona>]` (substituting the persona name with spaces, e.g.
  `[like a noir detective]`) and use vocabulary/phrasing matching that persona.
  `tts_instruct` should describe the persona's vocal style (cadence, register,
  mannerisms).

- A5_style_accent: the sub_axis is either `style=<X>` or `accent=<X>`.
  * For style=<X>: bake the style INTO the words (e.g. for pirate_speak,
    write "Arr, matey..."; for poetic, use poetic phrasing; for whispered,
    keep sentences short and intimate). `tts_instruct` should match the
    style ("hushed, breathy, intimate" for whispered, etc.).
  * For accent=<X>: keep content plain English but `tts_instruct` must
    explicitly name the accent ("Spoken with a clear British Received
    Pronunciation accent, crisp consonants and rounded vowels.").

- A6_speed_length: the sub_axis is either `speed=<X>` or `length=<X>`.
  * For speed=<X>: prepend the bracket tag (`[slow]`, `[fast]`, `[very fast]`,
    `[very slow]`) to the transcript. `tts_instruct` should describe the pace
    ("Rapid, breathless rattle of words" / "Measured, unhurried, room to
    breathe").
  * For length=<X>: write the response at the requested length (very_brief
    = 1 short sentence; very_detailed = ~5-6 sentences). No bracket tag.
    `tts_instruct` can stay neutral ("Natural pace, conversational.").

Length: aim for ~3-6 short sentences (target audio ~15-20s), shorter for
length=very_brief / length=brief, longer for length=detailed / very_detailed.

Return STRICT JSON:
{
  "transcript": "...spoken response, including any required bracket prefix...",
  "tts_instruct": "...natural-language director's note..."
}
\end{lstlisting}
\begin{lstlisting}[style=steerprompt,basicstyle=\ttfamily\fontsize{8}{9}\selectfont,literate={—}{{\textemdash}}1]
Sample: {sample_id}
Split: {split}
sub_axis: {sub_axis}
sub_axis_group: {sub_axis_group}

User utterance:
{user_utterance}

Rubric items the response must satisfy (these tell you what behaviour the
judge will look for):
{rubrics_block}

Generate the reference script now.
\end{lstlisting}

\subsection{Synthetic Reference-Audio Template}
The returned delivery instructions fill \texttt{sample\_context}, and the reference transcript, after removing parenthetical non-speech cues and normalizing whitespace, fills \texttt{transcript}. The audio request specifies PCM16 output. Generation metadata retain the transcript, delivery instructions, voice, model identifiers, and sample rate.
\begin{lstlisting}[style=steerprompt,basicstyle=\ttfamily\fontsize{8}{9}\selectfont,literate={—}{{\textemdash}}1,emptylines=0]
Read the following transcript based on the audio profile and director's note.

# Audio Profile
A helpful and professional personal assistant.

# Director's note
Style: Empathetic. Pace: Natural. Accent: American (Gen).

## Scene:
A quiet, professional remote workspace.

## Sample Context:
{sample_context}

## Transcript:
{transcript}
\end{lstlisting}

\end{document}